\pdfoutput=1
\documentclass{article}
\usepackage[main,final]{neurips_2026}

\newif\ifarxiv
\arxivfalse

\usepackage[utf8]{inputenc}
\usepackage[T1]{fontenc}
\usepackage{hyperref}
\usepackage{url}
\usepackage{booktabs}
\usepackage{array}
\usepackage{amsfonts}
\usepackage{nicefrac}
\usepackage{microtype}
\usepackage{xcolor}

\usepackage{graphicx}
\usepackage{subcaption}
\usepackage{enumerate}
\usepackage{amsmath}
\usepackage{amssymb}
\usepackage{mathtools}
\usepackage{amsthm}
\usepackage{enumitem}
\usepackage{multirow}
\usepackage[most]{tcolorbox}
\usepackage{tikz}
\usetikzlibrary{positioning, arrows.meta, decorations.pathreplacing, calc, shapes.geometric, fit, backgrounds}

\usepackage{algorithm}
\usepackage{algorithmic}

\usepackage[capitalize,noabbrev]{cleveref}
\usepackage[disable,textsize=tiny]{todonotes}

\makeatletter
\renewcommand{\maketitle}{%
  \par
  \begingroup
    \renewcommand{\thefootnote}{\fnsymbol{footnote}}
    \renewcommand{\@makefnmark}{\hbox to \z@{$^{\@thefnmark}$\hss}}
    \long\def\@makefntext##1{%
      \parindent 1em\noindent
      \hbox to 1.8em{\hss $\m@th ^{\@thefnmark}$}##1
    }
    \thispagestyle{empty}
    \@maketitle
    \@thanks
    \@notice
  \endgroup
  \let\maketitle\relax
  \let\thanks\relax
}
\makeatother

\newif\ifmarkrev
\markrevtrue
\newcommand{\rev}[1]{\ifmarkrev{\color{blue}#1}\else{#1}\fi}
\newcommand{\revcolor}{\ifmarkrev\color{blue}\fi}
\newif\ifmarkflag
\markflagtrue

\newcommand{\codeurl}{https://github.com/ZhihongCui/REINS}

\theoremstyle{plain}
\newtheorem{theorem}{Theorem}[section]

\theoremstyle{definition}
\newtheorem{definition}[theorem]{Definition}

\theoremstyle{remark}

\newcommand{\appref}[1]{Appendix~\ref{#1}}

\markrevfalse
\markflagfalse

\title{REINS: A Self-Evolving Agent Harness for Real-Time Trajectory Planning}

\author{%
  Zhihong Cui$^{1*}$ \quad Hengyu Liu$^{2*\dagger}$ \quad Haoran Tang$^{3}$ \quad Shijun Liu$^{4}$ \\
  \bfseries Amir Taherkordi$^{1}$ \quad Tor Skeie$^{1,5}$ \\
  \normalfont\small $^{1}$Department of Informatics, University of Oslo, Norway \\
  \normalfont\small $^{2}$Department of Computer Science, Aalborg University, Denmark \\
  \normalfont\small $^{3}$Department of Computing, The Hong Kong Polytechnic University, Hong Kong SAR, China \\
  \normalfont\small $^{4}$Shandong Key Laboratory of Foundational Software, School of Software, Shandong University, China \\
  \normalfont\small $^{5}$Simula Research Laboratory, Norway
}

\begin{document}

\maketitle
{\renewcommand{\thefootnote}{}\footnotetext{$^{*}$Equal contribution. \quad $^{\dagger}$Corresponding author: \texttt{heli@cs.aau.dk}}}

\begin{abstract}
Agent harnesses---runtime scaffolds that wrap an LLM's reasoning loop with tool dispatch, verification, and persistent memory---have become the standard substrate for deploying LLMs as autonomous agents, as demonstrated by Claude Code and Codex in software engineering. In trajectory planning, however, this paradigm has not taken hold: existing LLM-based trajectory planning methods either keep the LLM in the decision loop, exceeding the millisecond-level control budget, or offload it to offline rules or training-time supervision---leaving the runtime planner non-evolving. Intermediate hybrids still lack the verification-and-consolidation loop that makes harnesses reliable. We ask: \emph{what runtime harness would let an LLM planner meet real-time constraints while self-evolving into a deployable one?} We propose \textbf{REINS}, a harness built on three LLM--planner couplings: LLM outputs are (i) \textbf{grounded} to a learned variable-level dynamics model, (ii) \textbf{verified} via calibrated predicates and forward rollout, and (iii) \textbf{consolidated} into an indexed rule memory that serves future scenes in $\mathcal{O}(\log n)$ without re-invoking the LLM. The harness closes a self-evolution loop: familiar scenes resolve by memory lookup; novel scenes invoke the LLM, whose verified outputs enrich the memory. Without modifying LLM weights, REINS attains 98\%+ compliance, nearly 50\% collision reduction, and 1.9--4.6~ms latency on simulated and real-world benchmarks at the tactical decision layer, with LLM invocation rate monotonically declining as memory matures. Code: \url{\codeurl}.
\end{abstract}

\vspace{-2mm}
\section{Introduction}
\vspace{-2mm}
\begin{figure}[t]
    \centering
    \includegraphics[width=0.72\linewidth]{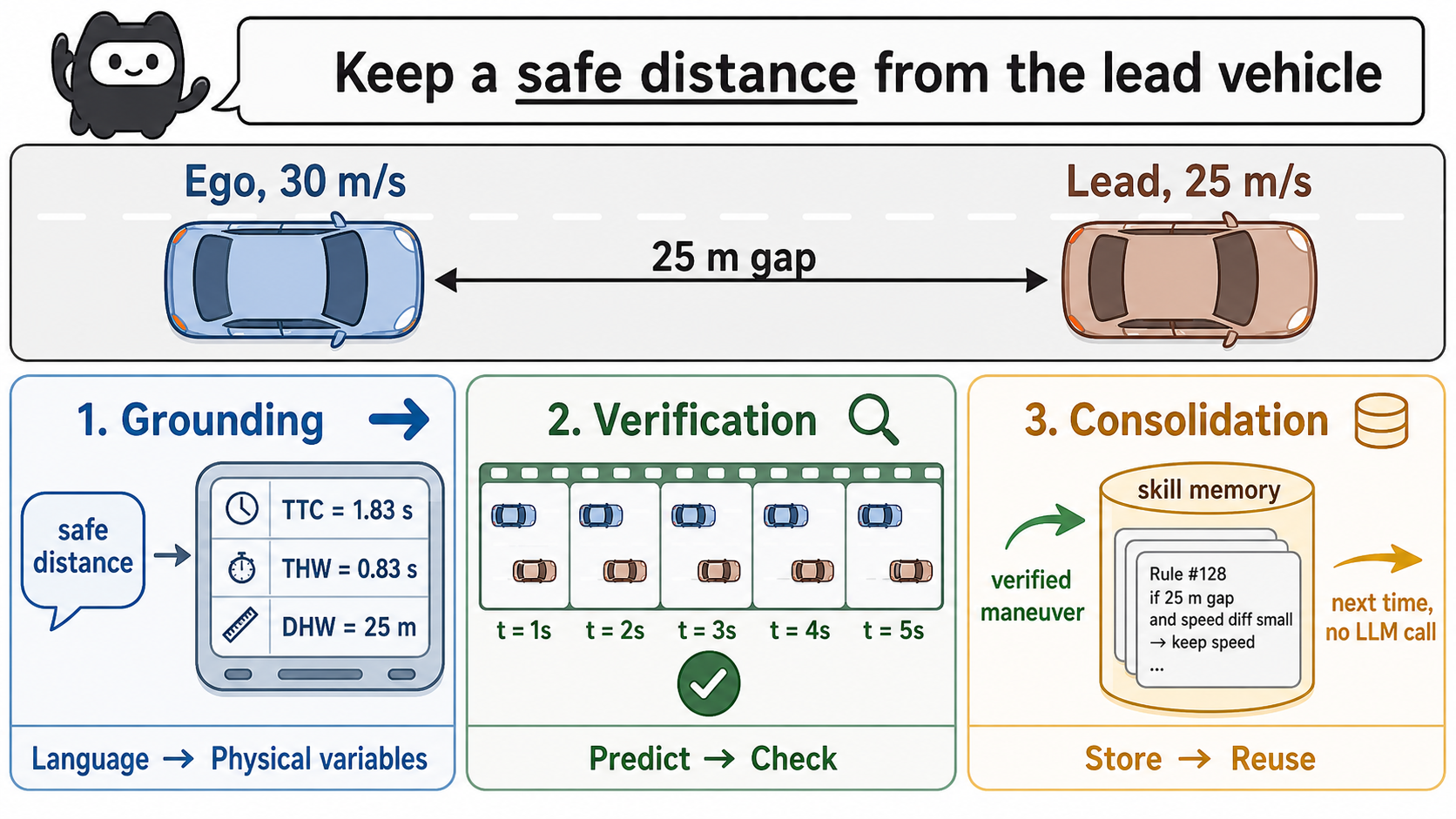}

    \caption{Three couplings an agent harness must enforce on the LLM planner, illustrated on a car-following scene with a 25-m gap. The LLM utterance ``keep a safe distance'' is (1) \textbf{grounded} to physical state variables (TTC, THW, DHW), (2) \textbf{verified} by forward rollout over a multi-step horizon, and (3) \textbf{consolidated} as an indexed rule keyed by these variables for instant reuse.}
    \vspace{-5mm}
    \label{fig:gapexample}
\end{figure}

Trajectory planning is the safety-critical core of autonomous driving: each decision must respect traffic rules, anticipate other agents, and complete within a millisecond-level control budget (e.g., 50~ms at 20~Hz)~\citep{zhang2021trajectory, falcone2007predictive, hubmann2016decentralized, li2024womd, montgomery1983decision}. Recent work shows that large language models (LLMs) bring rich semantic priors over traffic norms, agent intent, and safety reasoning that classical planners lack~\citep{manas2025integrating, fu2024dilu, mao2024agent, sha2023languagempc, cai2024driving, li2024driving}, yet integrating their reasoning into the runtime loop is bottlenecked: each LLM query takes seconds---orders of magnitude over the control budget---and naive integration leaves the planner either too slow to act or unable to adapt as deployment proceeds.

This is precisely the kind of deployment problem---running a stochastic LLM inside a low-latency, safety-sensitive loop---that the \textbf{agent harness} paradigm has recently addressed in software engineering. An agent harness wraps the LLM with a thin runtime layer of tool dispatch, intermediate-result verification, persistent memory, and safety enforcement---as exemplified by systems like Claude Code, Codex, and Cursor-Agent. On Terminal-Bench~\citep{merrill2026terminalbench}, harness design has emerged as a primary driver of ranking, with automatic harness optimization surpassing hand-engineered baselines~\citep{lee2026metaharness}. Trajectory planning, however, is a sharp test of the paradigm: its millisecond-level control budget is much tighter than the seconds-to-minutes scale at which coding harnesses operate, and the safety stakes are higher. In this regime, the harness's generic functions specialize into three couplings between LLM and planner, illustrated in Figure~\ref{fig:gapexample}: when an LLM suggests ``keep a safe distance'' in a car-following scene, the runtime must (i) \emph{ground} the suggestion to physical state variables (TTC, THW, DHW), (ii) \emph{verify} the resulting maneuver via forward rollout over a multi-step horizon, and (iii) \emph{consolidate} the verified maneuver into an indexed rule memory.

Existing LLM-based trajectory planning methods span two extremes plus intermediate hybrids, none of which constitutes a harness in the above sense: \emph{runtime invocation} methods keep the LLM in the decision loop but exceed the millisecond-level budget by orders of magnitude~\citep{fu2024dilu, mao2024agent, sha2023languagempc}; \emph{offline extraction} methods discard the LLM at runtime and freeze the planner against scenarios the offline corpus did not anticipate~\citep{li2024driving, tang2025plan}; and \emph{hybrid} designs partially orchestrate the LLM at runtime but stop short of a full harness, lacking explicit verification of LLM commitments and consolidation of verified ones into reusable structure~\citep{cui2026ctrail, zhang2025lead}. Across all three, the LLM's reasoning is either too slow to act on or too detached from the loop to evolve with deployment.

This raises our central question: \emph{what runtime harness would let an LLM planner satisfy real-time constraints while self-evolving into a deployable one?} Each of the three couplings is non-trivial under real-time control: grounding requires bridging natural-language commitments to the planner's threshold semantics~\citep{wang2025founder, zhai2025world}; verification must catch stochastic, occasionally unsafe LLM outputs before they reach control~\citep{manas2025integrating, zeng2025adrd}; and consolidation must escape per-call LLM latency~\citep{fu2024dilu, mao2024agent} via a memory pattern emerging in agentic AI~\citep{wang2023voyager, zhao2024expel, ma2024eureka} but not yet transferred to real-time planning. The three are also jointly necessary: without grounding, verification has nothing to check; without verification, consolidation stores unverified content; without consolidation, every scene re-pays the LLM latency cost. When all three are addressed, a \emph{self-evolution} loop emerges: familiar scenes resolve from the consolidated memory in $\mathcal{O}(\log n)$ time, novel scenes invoke the LLM, whose verified outputs enrich the memory, and reliance on the LLM decreases monotonically as memory matures, with no LLM weight update.

We propose \textbf{REINS}, a concrete realization of this harness pattern, instantiated for trajectory planning in driving, that addresses each coupling as follows.
\textbf{Grounding.} We learn a \emph{world dynamics model} over a factored variable space, fusing LLM-derived semantic priors with data-driven dependency discovery. Group-weighted ridge regression selects each variable's parents, and a per-variable Transformer predicts its evolution under the discovered graph. Every LLM output can then be projected onto this variable space and rolled out under the learned dynamics.
\textbf{Verification.} We compile each LLM-proposed maneuver into \emph{calibrated predicates} with thresholds derived from data (z-score, quantile, or absolute). Predicates are evaluated against the rollout produced by the dynamics model, and only commitments whose predicted trajectory satisfies all guards are released to control. Hallucinated or unsafe commitments are filtered before any maneuver is issued.
\textbf{Consolidation.} We persist verified commitments in an \emph{indexed rule memory} backed by a graph database, keyed by the grounding variables that triggered them. At runtime, parameterized graph queries retrieve scene-relevant rules in $\mathcal{O}(\log n)$ time without any LLM call. Recurring scenes are thus served by memory lookup, and the LLM is invoked only on a miss. Together, the three close the self-evolution loop: at each scene the planner first consults the memory; on a miss the LLM is invoked, its output grounded against the dynamics model, verified by rollout, and---if it passes---committed to memory for the next encounter. We provide an empirical demonstration that, under such a harness, an LLM planner monotonically reduces its own LLM-call rate while preserving safety and compliance. Our contributions can be summarized as:
\begin{itemize}[itemsep=1pt, leftmargin=10pt]
\item We extend the agent-harness paradigm to real-time trajectory planning, identifying \emph{three couplings}---grounding, verification, consolidation---any such harness must enforce, and instantiate it for driving; only the variable schema and the action space are domain-specific (\appref{app:scope}).
\item We design \textbf{REINS}, a harness that instantiates the three couplings via a learned world dynamics model, calibrated predicates with rollout verification, and an indexed rule memory.
\item We provide the first empirical evidence of self-evolution in real-time LLM trajectory planning: LLM invocation rate decreases monotonically over deployment with no LLM weight update.
\item We evaluate REINS on simulated and real-world driving benchmarks at the tactical decision layer, achieving 98\%+ compliance, nearly 50\% collision reduction, and 1.9--4.6~ms planning latency---well within the real-time control budget.
\end{itemize}
\section{Related Work}
\label{sec:related_work}

\textbf{LLM-Based Trajectory Planning.}
Recent LLM-based trajectory planners fall along three lines, each realizing some couplings while leaving others open. \emph{Runtime invocation} methods keep the LLM in the decision loop. DiLu~\citep{fu2024dilu} pairs LLM reasoning with an experience-replay memory; Agent-Driver~\citep{mao2024agent} treats the LLM as an orchestrator over perception and planning tools; LanguageMPC~\citep{sha2023languagempc} feeds chain-of-thought outputs into MPC. \rev{These systems gain semantic flexibility, but each LLM query takes seconds, and any output that reaches the planner is implicitly trusted, lacking the calibrated checks proposed for stochastic LLM outputs in safety-critical settings~\citep{manas2025integrating, zeng2025adrd}.}

\emph{Offline extraction} methods sit at the opposite end. LLaDA~\citep{li2024driving} extracts location-specific rules from driver handbooks, DriReg~\citep{cai2024driving} retrieves regulatory text via RAG, and Plan-R1~\citep{tang2025plan} distills LLM-shaped rewards into a reinforcement-learning policy. \rev{Removing the LLM at runtime buys real-time inference, but the planner becomes static, unable to consolidate new commitments against learned driving dynamics~\citep{wang2025founder, zhai2025world} or to adapt to scenarios the offline corpus did not anticipate.}

\emph{Hybrid} designs partially orchestrate the LLM at runtime. C-TRAIL~\citep{cui2026ctrail} runs a closed-loop Recall--Plan--Update cycle with trust-weighted MCTS injection; LeAD~\citep{zhang2025lead} pairs an end-to-end planner with a redundant LLM reasoning track. \rev{These improve over either extreme but lack explicit verification of LLM commitments and consolidation of verified ones into reusable structure; none grounds, verifies, and consolidates LLM outputs jointly under a real-time harness.}

\textbf{Agent Harnesses for LLM Deployment.}
Outside trajectory planning, the deployment of LLMs into real-world tasks has converged on the \emph{agent harness} pattern: a thin runtime layer that wraps the LLM with tool dispatch, intermediate-result verification, persistent memory, and safety enforcement. \rev{In autonomous coding, systems such as Claude Code and Codex turn general-purpose LLMs into reliable software engineers, and harness design has become a primary driver of ranking on Terminal-Bench~\citep{merrill2026terminalbench, lee2026metaharness}.} In open-ended decision-making, Voyager~\citep{wang2023voyager} accumulates a skill library of executable code through curiosity-driven exploration; Eureka~\citep{ma2024eureka} iteratively refines reward functions by evolutionary search over LLM proposals; ExpeL~\citep{zhao2024expel} distills task-specific experience into a retrievable knowledge base. These harnesses share three structural elements with REINS---grounding into a domain primitive space, verification of LLM outputs, and persistent reuse of verified ones---but none operate under millisecond real-time budgets nor expose their LLMs to the safety-critical constraints.

\section{Preliminaries}
\label{sec:preliminary}

We formalize the harness's primitives: a factored variable space (Section~\ref{sec:env}); a world dynamics model for \emph{grounding} (Section~\ref{sec:world_dynamics}); an executable rule representation for \emph{verification} and \emph{consolidation} (Section~\ref{sec:rule_repr}); and the problem definition (Section~\ref{sec:problem}). Full variable lists, template specifications, and the macro-group dependency enumeration appear in \appref{app:clear_details}.

\subsection{Driving Environment}
\label{sec:env}

\begin{definition}[Factored Variable Structure]
\label{def:variable_structure}
A \emph{factored variable structure} is $(\mathcal{X}, \mathcal{H}, \mathcal{E}_{\mathcal{G}}, \mathcal{E}_{\mathcal{H}})$: $\mathcal{X} = \{X_1, \ldots, X_d\}$ are observable variables (e.g., TTC, THW, DHW, ego\_speed, lead\_speed, road\_friction); $\mathcal{H} = \{\textsc{Env}, \textsc{Traffic}, \textsc{Ego}, \textsc{Action}, \textsc{Outcome}\}$ groups them by semantic role~\citep{chen2025causal}; $\mathcal{E}_{\mathcal{G}} \subseteq \mathcal{X} \times \mathcal{H}$ assigns each variable to a group; and $\mathcal{E}_{\mathcal{H}} \subseteq \mathcal{H} \times \mathcal{H}$ specifies permissible group-level dependencies (Eq.~\ref{eq:valid_edges} in \appref{app:variables}).
\end{definition}

\begin{definition}[State and Action Spaces]
\label{def:action}
The \emph{state} $s_t \in \mathcal{S} = \prod_i \mathrm{dom}(X_i)$ is a complete variable assignment. The \emph{action space} $\mathcal{A} = \{\texttt{LANE\_LEFT}, \texttt{LANE\_RIGHT}, \texttt{FASTER}, \texttt{SLOWER}, \texttt{IDLE}\}$ is a finite set of discrete meta-actions for the ego vehicle.
\end{definition}
\rev{The meta-actions are the interface of the tactical (behavior-planning) layer in the standard stack decomposition~\citep{paden2016survey}; continuous low-level controllers execute them, and all safety reasoning below (TTC, headway, accelerations) remains continuous.}

\subsection{World Dynamics Model}
\label{sec:world_dynamics}

\begin{definition}[Dependency Graph and Dynamics]
\label{def:dynamics}
A \emph{dependency graph} is a weighted DAG $\mathcal{G} = (\mathcal{X}, \mathcal{E}, w)$ with $w: \mathcal{E} \to \mathbb{R}$, where edges $(X_i, X_j) \in \mathcal{E}$ are admissible only if $(h(X_i), h(X_j)) \in \mathcal{E}_{\mathcal{H}}$~\citep{giamattei2024causality}. A \emph{dynamics function set} $\mathcal{F} = \{f_i\}_{i=1}^d$ predicts each variable from its parents over a temporal lag $L$~\citep{peters2017elements}:
\begin{equation}
\label{eq:dynamics}
\hat{X}_i^{t+1} = f_i\big(\mathrm{Pa}(X_i)^{t-L+1:t}\big), \quad \mathrm{Pa}(X_i) = \{X_j \mid (X_j, X_i) \in \mathcal{E}\}.
\end{equation}
An $H$-step rollout from $s_t$ under $(a_t,\ldots,a_{t+H-1})$ yields $\hat{\xi} = (\hat{s}_{t+1},\ldots,\hat{s}_{t+H})$.
\end{definition}

\subsection{Rule Representation}
\label{sec:rule_repr}

\begin{definition}[Predicate]
\label{def:predicate}
A \emph{predicate} is a Boolean function $p(s) = \mathbb{I}[s[X_i] \prec \tau]$ with comparison operator $\prec \in \{<, >, \leq, \geq\}$ and threshold $\tau \in \mathbb{R}$.
\end{definition}
\rev{TTC and THW guards are the surrogate safety measures of traffic engineering~\citep{gettman2003surrogate}; their calibrated thresholds act as RSS-style safe-distance bounds~\citep{shalev2017formal}.}

\begin{definition}[Rule Template]
\label{def:rule_template}
A \emph{rule template} $T$ specifies how predicates compose~\citep{manas2024tr2mtl}. We use seven templates: \texttt{IF-THEN}, \texttt{AND-THEN}, \texttt{THRESHOLD}, \texttt{PRIORITY}, \texttt{MULTI-ACTION}, \texttt{WEIGHTED-ACTION}, and \texttt{TEMPORAL-CHAIN} (specifications and examples in \appref{app:rules}).
\end{definition}

\begin{definition}[Executable Rule]
\label{def:rule}
An \emph{executable rule} is a tuple $r = (T, \mathcal{P}_r, a_r, c_r)$ with template $T$, guard predicates $\mathcal{P}_r$, recommended action $a_r \in \mathcal{A}$, and confidence $c_r \in [0,1]$. Activation requires all guards to hold:
\begin{equation}
\label{eq:rule_exec}
\mathrm{Exec}(r, s_t) = \bigwedge_{p \in \mathcal{P}_r} p(s_t).
\end{equation}
\end{definition}

\subsection{Problem Definition}
\label{sec:problem}

\begin{definition}[Self-Evolving Agent Harness for Trajectory Planning]
\label{def:problem}
\label{def:harness}
Let $\mathcal{D} = \{(s_t, a_t, s_{t+1})\}$ be a driving dataset and $\mathcal{M}$ a pre-trained LLM. The harness problem is to construct $\langle \hat{\mathcal{G}}, \hat{\mathcal{F}}, \mathcal{R}, \pi \rangle$ where $\hat{\mathcal{G}}, \hat{\mathcal{F}}$ are learned from $\mathcal{D}$ with priors from $\mathcal{M}$ (Def.~\ref{def:dynamics}), $\mathcal{R}$ is an indexed rule memory compiled offline from $\mathcal{M}$ (Def.~\ref{def:rule}), and $\pi: \mathcal{S} \to \mathcal{A}$ is a policy that, at each $s_t$, retrieves $\mathcal{R}_t = \mathrm{Retrieve}(s_t, \mathcal{R}) \subseteq \mathcal{R}$ and proceeds as follows:
\begin{itemize}[nosep, leftmargin=*]
    \item (\emph{hit}) if $\mathcal{R}_t \neq \emptyset$, $\pi$ scores actions over $\mathcal{R}_t$ and verifies them by rollout under $\hat{\mathcal{F}}$ within a real-time budget;
    \item (\emph{miss}) otherwise, $\mathcal{M}$ proposes a candidate commitment, which is grounded via $(\hat{\mathcal{G}}, \hat{\mathcal{F}})$ and verified by rollout; if all predicates pass, the commitment is compiled into a rule $r$ and $\mathcal{R} \leftarrow \mathcal{R} \cup \{r\}$.
\end{itemize}
The parameters of $\mathcal{M}$ remain unchanged, and $\mathcal{R}$ is monotone non-decreasing; under stationarity of the scene distribution, the per-scene LLM invocation rate $\Pr[\mathcal{R}_t = \emptyset]$ is non-increasing.
\end{definition}

\section{Methodology}

\begin{figure*}[t]
    \centering
    \includegraphics[width=0.86\linewidth]{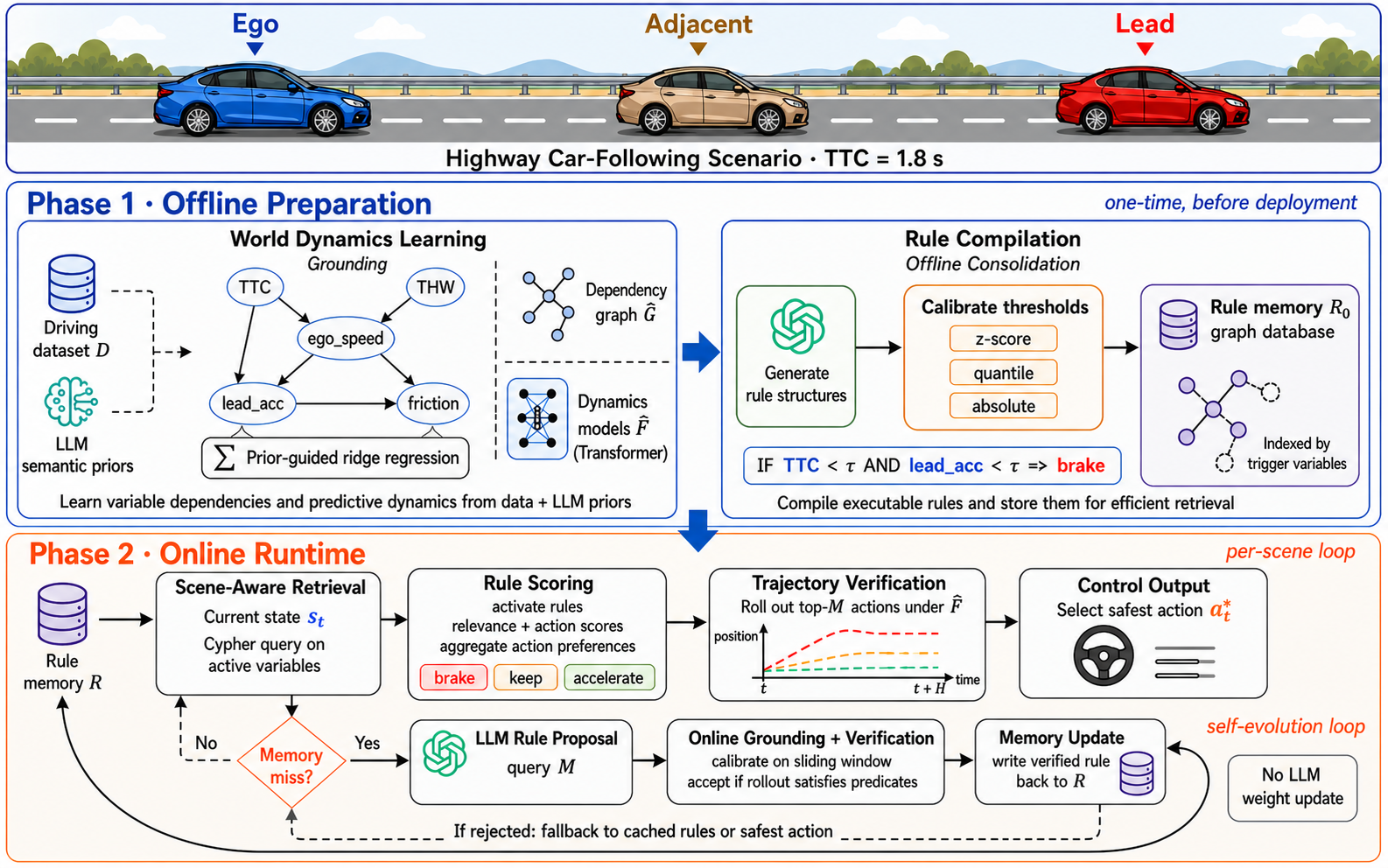}
    \caption{The REINS framework. \emph{Phase 1 (offline)}: learn dynamics $\hat{\mathcal{G}}, \hat{\mathcal{F}}$ and compile rules into initial memory $\mathcal{R}_0$. \emph{Phase 2 (online)}: per-scene retrieval and rollout-based verification produce control; on a miss, the LLM proposes a rule that is grounded, verified, and written back to $\mathcal{R}$, closing the self-evolution loop without any LLM weight update.}
    \vspace{-5mm}
    \label{fig:framework}
\end{figure*}

REINS realizes the harness as a two-phase system (Figure~\ref{fig:framework}). \emph{Offline Preparation} (Section~\ref{sec:offline}) prepares the world dynamics $\hat{\mathcal{G}}, \hat{\mathcal{F}}$ for \emph{grounding} and an initial rule memory $\mathcal{R}_0$ for offline \emph{consolidation}. \emph{Online Runtime} (Section~\ref{sec:online}) executes a per-scene loop: rule-guided planning performs retrieval, scoring, and \emph{verification} over $\mathcal{R}$, and online rule acquisition handles memory misses by querying $\mathcal{M}$ and writing verified rules back to $\mathcal{R}$, closing the \emph{self-evolution loop}.

\subsection{Offline Preparation}
\label{sec:offline}

\subsubsection{World Dynamics Learning}
\label{sec:world_dynamics_learning}

This module realizes \emph{grounding}: we jointly learn dependency graph $\hat{\mathcal{G}}$ and dynamics functions $\hat{\mathcal{F}}$ by combining LLM semantic priors with data-driven structure discovery. \rev{The full derivation and the dynamics loss are in \appref{app:world_dynamics_details}.}

\textbf{Dependency Graph.} The LLM provides edge confidence $\pi_{ij} \in [0,1]$ for each candidate edge $X_j \to X_i$ (\appref{app:prompts}); combined with macro-group constraints (Eq.~\ref{eq:valid_edges}), the unified prior is $\tilde{\pi}_{ij} = \alpha\,\pi_{ij} + (1-\alpha)\,\mathbb{I}\{(h(X_j), h(X_i)) \in \mathcal{E}_{\mathcal{H}}\}$. For each $X_i$, prior-guided ridge regression~\citep{hoerl1970ridge} predicts its value from lagged neighbors with adaptive penalty $\lambda_{ij} = \lambda_0(1-\tilde{\pi}_{ij})^\rho$, so high-prior edges are weakly regularized:
\begin{equation}
\hat{\mathbf{w}}_i = \arg\min_{\mathbf{w}} \sum_{t=L}^N \Big(X_i^t - \textstyle\sum_{j\neq i,\ell} w_{j\ell} X_j^{t-\ell}\Big)^2 + \sum_{j\neq i}\lambda_{ij}\|\mathbf{w}_j\|_2^2.
\label{eq:ridge}
\end{equation}
$\mathrm{Pa}(X_i)$ retains the top-$K$ variables by $\sum_\ell |\hat{w}_{j\ell}|$, yielding $\hat{\mathcal{E}}$ and weights $\hat{w}(X_j, X_i) = \sum_\ell |\hat{w}_{j\ell}|$.
% (Eq.~\ref{eq:graph_construct}).

\textbf{Dynamics Functions.} Given $\hat{\mathcal{G}}$, each $f_i$ is a multi-head Transformer~\citep{vaswani2017attention} trained to predict $X_i^{t+1}$ from $\mathrm{Pa}(X_i)^{t-L+1:t}$ via mean squared error (loss form in Eq.~\ref{eq:dyn_loss}). Training curves and hyperparameters are in \appref{app:training_loss}.

\subsubsection{Rule Compilation}
\label{sec:rule_compilation}

This module realizes offline \emph{consolidation}: compiling LLM knowledge into executable rules. Full equations and per-step prose are in \appref{app:rule_compilation_details}.

\textbf{Rule Structure Generation.} For each edge $(X_j, X_i) \in \hat{\mathcal{E}}$, the LLM generates a rule $r = (T, \mathcal{P}_r, a_r, c_r)$ (Eq.~\ref{eq:rule_gen}) with template $T$ (Def.~\ref{def:rule_template}), guard predicates $\mathcal{P}_r$ with symbolic thresholds, recommended action $a_r \in \mathcal{A}$, and confidence $c_r$ derived from edge weight $\hat{w}(X_j, X_i)$ (prompts in \appref{app:prompts}).

\textbf{Threshold Calibration.} Each predicate's threshold $\tau_i$ is calibrated from $\mathcal{D}$ in three forms~\citep{luo2024graph}---z-score $\mu_i + \zeta\sigma_i$, quantile $Q_i(\zeta)$, or absolute constant $\tau_0$---and the comparison operator $\prec$ is selected to maximize separability of safety outcome $y \in \{0,1\}$:
\begin{equation}
\prec^* = \arg\max_{\prec \in \{<, >\}}
\big|\mathbb{E}[y \mid X_i \prec \tau_i] - \mathbb{E}[y \mid X_i \not\prec \tau_i]\big|.
\end{equation}
Rule activation then follows Eq.~\ref{eq:rule_exec}.

\textbf{Graph Database Storage.} Compiled rules are stored in a Neo4j graph database~\citep{webber2012programmatic} with rule nodes connected to variable nodes via \texttt{TRIGGERS} (input) and \texttt{AFFECTS} (output) edges, enabling $O(\log|\mathcal{R}|)$ retrieval. The full schema (Figure~\ref{fig:rule_database_schema}) is in \appref{app:rule_compilation_details}; database statistics are in \appref{app:rule_database_statistics}.

\subsection{Online Runtime}
\label{sec:online}

\subsubsection{Rule-Guided Planning}
\label{sec:rule_inference}

This module realizes the runtime planning loop of Def.~\ref{def:harness}---retrieval, scoring, and \emph{verification} \rev{(details in \appref{app:planning_details})}.

\textbf{Scene-Aware Retrieval.} Given $s_t$, active variables $\{X_i\}$ trigger parameterized Cypher queries~\citep{francis2018cypher} generated by the LLM (read-only, schema-constrained; \appref{app:prompts}). The retrieval map $G: (\{X_i\}, K) \mapsto \mathcal{R}_t \subseteq \mathcal{R}$ (Eq.~\ref{eq:cypher_gen}) returns up to $K$ rules ranked by confidence, restricted to triggers whose dependency direction respects the macro-group constraints.

\textbf{Action Scoring.} For each $r \in \mathcal{R}_t$ activated via Eq.~\ref{eq:rule_exec}, a Transformer-based scorer~\citep{liang2022trajformer} produces relevance $S_{\text{rel}}(r, s_t) \in [0,1]$ and action preference $S_{\text{act}}(r, s_t, a)$, yielding combined score $f(r, a \mid s_t) = S_{\text{rel}} \cdot S_{\text{act}} \cdot c_r$ (Eqs.~\ref{eq:rule_score}--\ref{eq:rule_action_score}; training loss in Eq.~\ref{eq:scorer_loss}). Multiple rules recommending action $a$, denoted $\mathcal{R}_t(a) = \{r \in \mathcal{R}_t \mid a_r = a\}$, are aggregated by noisy-OR:
\begin{equation}
P(a \mid s_t) \;\propto\; 1 - \prod_{r \in \mathcal{R}_t(a)}\big(1 - f(r, a \mid s_t)\big).
\label{eq:prior}
\end{equation}
The full normalized form and argmax decoder are in \appref{app:planning_details}.

\textbf{Trajectory Verification.} The learned dynamics $\hat{\mathcal{F}}$ verifies safety via forward rollout. For the top-$M$ actions ranked by $P(a \mid s_t)$, we predict $H$-step trajectories $\hat{\xi}(a) = (\hat{s}_{t+1}, \ldots, \hat{s}_{t+H})$ via $\hat{s}_{t+h+1} = \Phi(\hat{s}_{t+h}, a; \hat{\mathcal{G}}, \hat{\mathcal{F}})$ (Eq.~\ref{eq:rollout}), and select $a^\star_t = \arg\max_{a \in \mathcal{A}_M} J(\hat{\xi}(a))$, where $J$ penalizes collisions, rule violations, and control effort (\appref{app:metrics}); the first control is executed in receding-horizon~\citep{zhang2021world}.

\subsubsection{Online Rule Acquisition}
\label{sec:rule_acquisition}

This module realizes Def.~\ref{def:harness}'s miss-handling path through four steps \rev{(full equations and the monotonicity argument in \appref{app:acquisition_details})}.

\textbf{Miss Detection.} We operationalize the miss criterion ($\mathcal{R}_t = \emptyset$) with an additional confidence guard against weakly supported matches:
\begin{equation}
\label{eq:miss}
\mathrm{Miss}(s_t) \;=\; \mathbb{I}\!\left[\mathcal{R}_t = \emptyset\right] \;\vee\; \mathbb{I}\!\left[\max_{r \in \mathcal{R}_t,\, a \in \mathcal{A}} f(r, a \mid s_t) < \eta\right],
\end{equation}
with tuned threshold $\eta \in (0, 1)$.

\textbf{Rule Proposal and Online Verification.} On a miss, $\mathcal{M}$ proposes $\tilde{r} = (\tilde{T}, \tilde{\mathcal{P}}, \tilde{a}, \tilde{c})$ (Eq.~\ref{eq:llm_proposal}) conforming to Defs.~\ref{def:predicate}--\ref{def:rule_template} (\appref{app:prompts}). Symbolic thresholds in $\tilde{\mathcal{P}}$ are calibrated on a sliding window $\mathcal{D}_t^W = \{s_{t-W+1}, \ldots, s_t\}$ so they track the current distribution. The harness then rolls out $\tilde{a}$ from $s_t$ under $\hat{\mathcal{F}}$ and accepts $\tilde{r}$ iff every predicate holds along the rollout:
\begin{equation}
\label{eq:online_verify}
\mathrm{Accept}(\tilde{r}, s_t) \;=\; \bigwedge_{p \in \tilde{\mathcal{P}}} \bigwedge_{h=1}^{H} p(\hat{s}_{t+h}).
\end{equation}
On rejection the planner falls back to $\arg\max_a P(a \mid s_t)$ over $\mathcal{R}_t$ if $\mathcal{R}_t \neq \emptyset$; otherwise the harness raises a human-takeover alert and executes the safest action under $\hat{\mathcal{F}}$ by maximizing minimum lookahead TTC (Eq.~\ref{eq:safe_action} in \appref{app:acquisition_details}).

\textbf{Memory Update.} Accepted $\tilde{r}$ is keyed by content hash $\kappa(\tilde{r}) := \kappa(\tilde{T}, \tilde{\mathcal{P}}, \tilde{a})$ and inserted into Neo4j; on collision, confidences are averaged instead of duplicating the node:
\begin{equation}
\label{eq:memory_update}
\mathcal{R} \;\longleftarrow\;
\begin{cases}
\mathcal{R} \cup \{\tilde{r}\} & \kappa(\tilde{r}) \notin \mathrm{Keys}(\mathcal{R}), \\[2pt]
\mathcal{R} \text{ with } c_{r'} \leftarrow \tfrac{1}{2}(c_{r'} + \tilde{c}) & \exists\, r' \in \mathcal{R}: \kappa(r') = \kappa(\tilde{r}).
\end{cases}
\end{equation}
The collision-branch rationale (averaging vs.\ duplication, exclusion of $\tilde{c}$ from $\kappa$) is in \appref{app:acquisition_details}.

\textbf{Self-Evolution Properties.} Eq.~\ref{eq:memory_update} structurally enforces $|\mathcal{R}_{t+1}| \geq |\mathcal{R}_t|$ (memory monotonicity) and $\theta_{\mathcal{M}} \equiv \theta_{\mathcal{M}}^{(0)}$ (no LLM update). Under stationarity of the scene distribution, the per-state miss rate is non-increasing as $\mathcal{R}$ accumulates coverage:
\begin{equation}
\label{eq:self_evolve_props}
|\mathcal{R}_{t+1}| \geq |\mathcal{R}_t|, \quad \Pr[\mathrm{Miss}(s_{t+1})] \leq \Pr[\mathrm{Miss}(s_t)], \quad \theta_{\mathcal{M}} \equiv \theta_{\mathcal{M}}^{(0)},
\end{equation}
so the per-scene LLM invocation rate decays monotonically with $\Pr[\mathrm{Miss}]$ (full argument in \appref{app:acquisition_details}; validated in Section~\ref{sec:self_evolution}).

\vspace{-1mm}
\section{Experiment}
\vspace{-1mm}

The main body reports overall comparison (Section~\ref{sec:perf_eval}), ablation (Section~\ref{sec:ablation}), self-evolution dynamics on Highway (Section~\ref{sec:self_evolution}), \rev{and a summary of the robustness, transfer, and scalability results. Appendices~\ref{app:additional} and~\ref{app:review} report trajectory accuracy, sensitivity, extended self-evolution traces, a case study, the rule database, and the fairness and robustness studies.}

\subsection{Experiment Setting}
\label{sec:exp_setting}
 \textbf{\rev{Datasets and Protocols.}} We evaluate REINS on four \texttt{highway-env}~\citep{highway-env} scenarios---Highway, Merge, Roundabout, Intersection (100 steps at 5\,Hz, 15 observable vehicles per episode)---and two drone-captured real-world datasets~\citep{krajewski2018highd,krajewski2020round}: highD (German highway, 4{,}237 vehicles, 578 lane changes) and rounD (German roundabouts, 2{,}156 vehicles, 423 yield/merge events). \rev{Surrounding vehicles in simulation are \emph{reactive} (IDM car-following~\citep{treiber2000congested} with MOBIL lane changes~\citep{kesting2007general}), so they brake, yield, and change lanes in response to the ego. The real-world datasets run in \emph{closed-loop replay}: one recorded vehicle is controlled by the planner while the others follow their recorded trajectories, so these rows reflect human behavior rather than a simulator's; a collision is registered when bounding boxes overlap or the ego leaves the drivable lane. A reactive-replay variant following the nuPlan protocol~\citep{caesar2021nuplan} is in \appref{app:reactive_noise}. All methods observe clean factored variables; perception, tracking, and map errors lie outside the evaluated layer (Section~\ref{sec:limitations}).}

 \textbf{Evaluation Metrics.} \rev{\textbf{Compl.} (Rule Compliance) is the fraction of steps at which all active rules hold, measured for each method against its own executable rule set; an audit of all methods against REINS's compiled set as a common reference leaves the ordering unchanged (\appref{app:fairness}). \textbf{RE} (Risk-aware Efficiency) is task reward minus a TTC-based risk penalty. \textbf{CR} (Collision Rate) is the fraction of \emph{episodes} with at least one contact among the 15 vehicles over 100 steps, an episode-level rate in deliberately stressed traffic. \textbf{PE} (Planning Efficiency) is the mean wall-clock ms of the synchronous action-selection computation per step for every method; for REINS it excludes the asynchronous rule-acquisition pipeline, profiled separately in \appref{app:miss_cycle}. TTC and THW, which underlie the guards and the RE penalty, are the surrogate safety measures of traffic engineering~\citep{gettman2003surrogate}, and Compl.\ and CR echo the rule-conformance and collision outcomes reported in scenario-based safety assessment~\citep{riedmaier2020survey}.} Results are averaged over five seeds; formal definitions in \appref{app:metrics}.

 \textbf{Baselines.} We compare with eight baselines covering the regimes from Section~\ref{sec:related_work} plus classical anchors:
(i) \emph{classical} (MPC~\citep{falcone2007predictive}, MCTS~\citep{hubmann2016decentralized});
(ii) \emph{runtime invocation} (DriveSpeak~\citep{cui2024drive});
(iii) \emph{offline extraction} (LLaDA~\citep{li2024driving}, DriReg~\citep{cai2024driving}, Plan-R1~\citep{tang2025plan});
(iv) \emph{symbolic/hybrid} (\rev{PDDL~\citep{guan2023leveraging,liu2023llmp}}, SymAgent~\citep{liu2025symagent}).
\rev{All methods share the observation space, action space, episodes, and LLM backbone (GPT-4o, temperature 0); descriptions, the shared and unshared conditions, and a matched re-run of PDDL and DriReg over REINS's vocabulary and calibration pipeline are in Appendices~\ref{app:baselines} and~\ref{app:fairness}.}

\begin{table*}[t]
\centering
\small
\setlength{\tabcolsep}{1pt}
\renewcommand{\arraystretch}{1.05}
\fontsize{8.4pt}{10pt}\selectfont
\begin{tabular}{c|c|c|cc|cc|cccc|c}
\toprule
\textbf{Dataset} & \textbf{Scenario} & \textbf{Metric}
    & \textbf{DriveSpeak}
    & \textbf{LLaDA}
    & \textbf{MPC}
    & \textbf{MCTS}
    & \textbf{PDDL}
    & \textbf{DriReg}
    & \textbf{SymAgent}
    & \textbf{Plan-R1}
    & \textbf{\textsf{REINS}}\\
\midrule
\multirow{16}{*}{\rotatebox{90}{\textit{Simulation}}}
& \multirow{4}{*}{Hig.}
& Compl. ($\uparrow$) &- & -& -& -& 0.9123 & 0.9452& 0.9561 & 0.9717& \textbf{0.9956}\\
&& RE ($\uparrow$) & -6.0360& -5.8537 & -5.5613& -5.2311 & -4.7868& -4.1282 & -3.8662 & -3.5623 & \textbf{0.5228}\\
&& CR ($\downarrow$,\%) & 38.28& 35.51 & 14.27 & 16.76& 25.93& 21.76 & 16.52& 12.83& \textbf{5.67}\\
&& PE ($\downarrow$, ms) & 5.8921 & 5.2995 & 2.4742& \textbf{2.3591}& 4.9521 & 4.3127& 3.9621 & 3.7391&2.4998\\
\cmidrule{2-12}
& \multirow{4}{*}{Mer.}
& Compl. ($\uparrow$)&- & -& -& - & 0.9011 & 0.9382 &0.9572 & 0.9641 & \textbf{0.9924} \\
&& RE ($\uparrow$) & -5.5730& -5.2621& -5.3891&  -5.1890& -4.7129 & -4.1067 & -3.5622 & -3.1023 & \textbf{-0.0400}\\
&& CR ($\downarrow$,\%) & 35.62 & 22.84 & 27.91& 23.64 & 21.52& 20.56 & 16.83 & 14.78& \textbf{6.86} \\
&& PE ($\downarrow$, ms) & 5.2120 & 4.3615 & 1.7354 & \textbf{1.6526}& 3.8549 & 2.8122& 2.5697 & 2.2418& 1.8928\\
\cmidrule{2-12}
& \multirow{4}{*}{Rou.}
& Compl. ($\uparrow$) & -& -&- & -& 0.8838 & 0.9129 & 0.9411& 0.9568 & \textbf{0.9925}\\
&& RE ($\uparrow$) & -7.6873 & -7.1069 & -6.7722& -6.5113 & -6.4895 & -6.3457& -6.2313 & -6.0256& \textbf{-0.5444}\\
&& CR ($\downarrow$,\%) & 39.19 & 36.63 & 24.27& 21.62& 23.84 & 25.13 & 24.74& 18.57& \textbf{9.01} \\
&& PE ($\downarrow$, ms) & 8.2673 & 7.6541 & 4.2134 & 4.0267 & 5.3793 & 5.1682 & 5.0713 & 4.8621 & \textbf{3.6243}\\
\cmidrule{2-12}
& \multirow{4}{*}{Int.}
& Compl. ($\uparrow$) & -& -& -& -& 0.8789 & 0.9153 & 0.9274 & 0.9553 &\textbf{0.9838}\\
&& RE ($\uparrow$) & -7.9740 & -7.5800 & -7.1345 &-7.0273 & -6.9792 & -6.7632 & -6.3314 & -6.0322 & \textbf{-0.5747} \\
&& CR ($\downarrow$,\%) & 44.27 & 42.22 & 25.38& 23.78 & 32.12 & 29.89 & 25.34 & 20.37 & \textbf{9.08}\\
&& PE ($\downarrow$, ms) & 10.9324& 9.3549 & 4.9653 & 4.7519& 6.7682& 6.2793& 5.9932 & 5.5733 & \textbf{4.5807} \\
\midrule
\multirow{8}{*}{\rotatebox{90}{\textit{Real-world}}}
& \multirow{4}{*}{highD}
& Compl. ($\uparrow$) & -& -& -& -& 0.9078 & 0.9289 & 0.9423 & 0.9612 &\textbf{0.9867}\\
&& RE ($\uparrow$) & -5.8234 & -5.5127 & -5.2891 & -5.0124 & -4.6523 & -4.2178 & -3.7856 & -3.4521 & \textbf{0.3847} \\
&& CR ($\downarrow$,\%) & 35.21 & 32.87 & 18.34 & 19.56 & 23.12 & 20.89 & 17.23 & 12.78 & \textbf{6.12}\\
&& PE ($\downarrow$, ms) & 6.1234 & 5.5672 & 2.6123 & \textbf{2.4891} & 5.1234 & 4.5623 & 4.1234 & 3.8912 & 2.6234 \\
\cmidrule{2-12}
& \multirow{4}{*}{rounD}
& Compl. ($\uparrow$) & -& -& -& -& 0.8956 & 0.9187 & 0.9378 & 0.9534 &\textbf{0.9856}\\
&& RE ($\uparrow$) & -7.4521 & -6.9834 & -6.5123 & -6.2478 & -6.1234 & -5.8912 & -5.6234 & -5.3178 & \textbf{-0.4123} \\
&& CR ($\downarrow$,\%) & 41.23 & 38.56 & 25.34 & 23.12 & 26.78 & 24.23 & 20.89 & 18.47 & \textbf{9.45}\\
&& PE ($\downarrow$, ms) & 8.5621 & 7.8923 & 4.3521 & 4.1234 & 5.6234 & 5.2178 & 4.8912 & 4.5623 & \textbf{3.8912} \\
\bottomrule
\end{tabular}%
\caption{Performance comparison across simulated scenarios and real-world datasets. Methods without explicit rule representations (MPC, MCTS, DriveSpeak, LLaDA) cannot report Compliance \rev{against their own rules} as they lack executable rule constraints\rev{; their compliance against REINS's compiled set as a common reference is given in Table~\ref{tab:common_compliance}}.}
\label{tab:mainresults}
\vspace{-6mm}
\end{table*}

\subsection{Performance Evaluation}
\label{sec:perf_eval}

Performance evaluation results are reported in Table~\ref{tab:mainresults}, with trajectory accuracy analysis in \appref{app:trajectory}.

\textbf{Effectiveness Analysis.} (i) {REINS} beats all baselines, improving over Plan-R1 by 2.4\%--3.6\% Compl., 3.06--5.48 RE, and 49\%--56\% CR reduction---validating that grounding, verification, and consolidation act jointly. (ii) All methods incur higher CR on complex scenarios (Roundabout, Intersection) than structured ones (Highway, Merge); {REINS} degrades least, owing to context-aware retrieval. (iii) On highD and rounD, {REINS} maintains Compl.\ above 0.98 with 6.6--9.0-point CR reduction over Plan-R1, transferring to naturalistic driving without domain-specific tuning. (iv) Executable rule methods (PDDL, DriReg, SymAgent, Plan-R1) beat language-guided ones (DriveSpeak, LLaDA, with 22\%--44\% CR for ungrounded rules), yet their static thresholds and uniform activation still trail {REINS}'s data-driven calibration and variable-indexed retrieval\rev{; a matched re-run over REINS's vocabulary and calibration narrows PDDL's and DriReg's CR by 3--5 points but leaves the gap intact (\appref{app:fairness})}. (v) Classical planners (MPC, MCTS) are fast but unsafe (CR 14\%--28\%); Plan-R1 is the best baseline but plateaus under uniform rule application.

\textbf{Efficiency Analysis.}
{REINS} completes planning in 1.9--4.6\,ms per step, up to 58\% faster than runtime LLM baselines (DriveSpeak, LLaDA) and 15--33\% faster than Plan-R1. Two factors contribute: (i) variable-indexed retrieval activates only 2--3 rules per step via $O(\log|\mathcal{R}|)$ queries, and (ii) learned dynamics enable closed-form rollout without iterative LLM calls. \rev{PDDL makes no runtime LLM call either, yet re-grounds every precondition at each step (3.71 of its 4.95\,ms on Highway; \appref{app:pddl_profile}), exactly the cost that indexed consolidation removes.}

\begin{table}[!tbp]
\centering
\setlength{\tabcolsep}{4pt}
\fontsize{8.4pt}{10pt}\selectfont
\begin{tabular}{c|c|cccc|c}
\toprule
    & \multirow{2}{*}{\textbf{Metric}} & \textbf{w/o} & \textbf{w/o LLM} & \textbf{w/o Rule} & \textbf{w/o} & \multirow{2}{*}{\textbf{Full}} \\
    &  & \textbf{World} & \textbf{Prior} & \textbf{Generation} & \textbf{Retrieval} &  \\
\midrule
\multirow{4}{*}{Hig.}
    & Compl. ($\uparrow$) & 0.9521 & 0.9628 & 0.9312 & 0.9845 & \textbf{0.9956}\\
    & RE ($\uparrow$)  & -1.9234 & -1.5390 & -2.4156 & -0.8921 & \textbf{0.5228} \\
    & CR ($\downarrow$,\%) & 8.47 & 8.26 & 11.23 & 8.12 & \textbf{5.67}\\
    & PE ($\downarrow$, ms) & 2.9532 & 2.6831 & 2.5124 & 3.8921 & \textbf{2.4998}\\
\midrule
\multirow{4}{*}{Mer.}
    & Compl. ($\uparrow$) & 0.9687 & 0.9738 & 0.9198 & 0.9812 & \textbf{0.9924}\\
    & RE ($\uparrow$) & -1.5621 &-1.2619 & -2.1893 & -0.7234 & \textbf{-0.0400}\\
    & CR ($\downarrow$,\%) & 9.23 & 9.08 & 12.87 & 8.89 & \textbf{6.86} \\
    & PE ($\downarrow$, ms) & 2.6234 & 2.1832 & 1.9021 & 2.9834 & \textbf{1.8928}\\
\midrule
\multirow{4}{*}{Rou.}
    & Compl. ($\uparrow$) & 0.9534 & 0.9611 & 0.9087 & 0.9756 & \textbf{0.9925}\\
    & RE ($\uparrow$) & -2.3456 & -1.1783 & -3.1234 & -1.2134 & \textbf{-0.5444}\\
    & CR ($\downarrow$,\%) & 12.34 & 11.84 & 15.89 & 11.23 & \textbf{9.01} \\
    & PE ($\downarrow$, ms) & 4.1245 & 3.8011 & 3.6521 & 5.2134 & \textbf{3.6243} \\
\midrule
\multirow{4}{*}{Int.}
    & Compl. ($\uparrow$) & 0.9456 & 0.9582 & 0.8923 & 0.9689 & \textbf{0.9838}\\
    & RE ($\uparrow$) & -2.5678 & -1.9812 & -3.4521 & -1.3456 & \textbf{-0.5747}\\
    & CR ($\downarrow$,\%) & 13.45 & 12.23 & 17.56 & 11.12 & \textbf{9.08}\\
    & PE ($\downarrow$, ms) & 4.9521 & 4.7214 & 4.6234 & 6.1234 & \textbf{4.5807}\\
\bottomrule
\end{tabular}
\caption{Ablation study across four driving scenarios.}
\vspace{-5mm}
\label{tab:ablation}
\end{table}

\subsection{Ablation Study}
\label{sec:ablation}
We conduct ablation experiments on all four simulated scenarios to evaluate component contributions. Table~\ref{tab:ablation} reports results for four variants:
(i) \textbf{w/o World Model}: replacing learned dynamics with constant-velocity propagation;
(ii) \textbf{w/o LLM Prior}: using purely data-driven structure learning without semantic guidance;
(iii) \textbf{w/o Rule Generation}: substituting LLM-generated rules with hand-crafted ones;
(iv) \textbf{w/o Retrieval}: applying all rules uniformly without context-aware selection.

Removing the world dynamics model causes the largest safety degradation (CR +49\%/+48\% on Highway/Intersection), confirming multi-step prediction is essential. Without LLM priors, CR rises 31--46\%, showing that semantic guidance regularizes dependency discovery beyond purely data-driven methods. Hand-crafted rules yield the largest compliance drop (0.9956$\to$0.9312 on Highway) and nearly double CR, exposing the limited coverage of manual engineering\rev{; even so, they remain safer inside the harness than every matched no-harness baseline (\appref{app:fairness})}. Disabling context-aware retrieval raises both CR (22--43\%) and PE (34--58\%) via conflicting constraints. The four roles are complementary: dynamics enables predictive verification, LLM priors regularize structure, rule generation ensures coverage, and selective retrieval prevents conflicts.

\subsection{Self-Evolution Analysis}
\label{sec:self_evolution}

\begin{table}[t]
\centering
\setlength{\tabcolsep}{4pt}
\fontsize{8.4pt}{10pt}\selectfont
\begin{tabular}{l|c|c|cc}
\toprule
\textbf{Checkpoint} & $|\mathcal{R}|$ & \textbf{LLM Call Rate (\%)} & \textbf{Compl.} ($\uparrow$) & \textbf{CR (\%)} ($\downarrow$) \\
\midrule
ep.\ 250                                  & 113 & 70.0  & 0.9885 & 7.05 \\
ep.\ 500                                  & 127 & 48.44 & 0.9920 & 6.11 \\
ep.\ 750                                  & 138 & 25.0  & 0.9942 & 5.83 \\
ep.\ 1000 (final)                         & \textbf{145} & \textbf{12.81} & \textbf{0.9956} & \textbf{5.67} \\
\midrule
\textit{Frozen $\mathcal{R}_0$} (1000-ep eval) & 88 & --- & 0.9800 & 9.63 \\
\bottomrule
\end{tabular}
\caption{Self-evolution dynamics on Highway over 1000 episodes\rev{; LLM call rate is the miss rate (Eq.~\ref{eq:miss}) in the 250-episode window ending at each checkpoint}.}
\vspace{-5mm}
\label{tab:self_evolution}
\end{table}

\rev{We deploy REINS for 1000 episodes on Highway from the offline-compiled memory $\mathcal{R}_0$ ($|\mathcal{R}_0|{=}88$), reporting four checkpoint snapshots alongside a \emph{Frozen $\mathcal{R}_0$} control that runs the same window with acquisition disabled (Table~\ref{tab:self_evolution}; other scenarios in \appref{app:self_evolution_extended}).} Across deployment, $|\mathcal{R}|$ grows from 88 to 145 (saturating after ep.\ 500; Eq.~\ref{eq:self_evolve_props}), the LLM call rate decays from 70.0\% to 12.81\% (a 5.5$\times$ reduction with no weight update), Compl.\ rises from 0.9885 to 0.9956, and CR drops from 7.05\% to 5.67\%. Against the Frozen $\mathcal{R}_0$ control over the same horizon (0.9800 Compl., 9.63\% CR), online acquisition adds $+0.0156$ Compl.\ and removes 3.96 CR points---the headline self-evolution gain. Mapped to the three couplings: monotone $|\mathcal{R}|$ under $\kappa$-hash deduplication validates consolidation, the joint Compl.\,$\uparrow$/CR\,$\downarrow$ trajectory validates rollout-based verification, and the 5.5$\times$ call-rate drop validates the grounding--consolidation loop, as variable-keyed memory hits replace LLM calls on similar scenes.

\ifarxiv
\begingroup\revcolor
\textbf{What verification filters.} The call rates above imply roughly 39{,}000 LLM proposals over the 100{,}000 deployment steps. Of these, 2.1\% failed schema validation, 18.0\% were rejected by rollout verification, and 79.9\% were accepted; acceptances overwhelmingly hash-collide with rules already in memory and only update confidences, deduplicating to the 57 unique rules that grow $\mathcal{R}$ from 88 to 145. A manual audit judged 54 of the 57 consistent with domain guidelines; the three remaining were over-specific and never re-activated, and episodes in which an acquired rule fired show no elevated CR against episodes in which none fired (5.4\% vs.\ 5.8\%; \appref{app:verification_residual}).
\endgroup
\else
\rev{Of the roughly 39{,}000 proposals implied by these call rates, 2.1\% failed schema validation, 18.0\% were rejected by rollout verification, and the accepted remainder deduplicated to 57 unique rules, 54 of which an audit judged consistent with domain guidelines (\appref{app:verification_residual}).}
\fi

\ifarxiv
\else
\begingroup\revcolor
\textbf{Robustness, Transfer, and Scalability.}
\label{sec:robustness}
\appref{app:review} stresses these results. Rules regenerated with Claude Sonnet 4.5 or Qwen2.5-72B-Instruct keep Highway CR at 5.89\% and 6.24\% (vs.\ 5.67\%), and prompt paraphrases span 5.58--5.91\%. Under 2.5--10\% observation noise and 5\% frame drops, REINS CR rises to at most 8.14\% while Plan-R1 rises to 19.87\%. A memory compiled only from highD/rounD starts at an 86.2\% call rate on highway-env and recovers to within 0.5--1.1 CR points of the native memory within 1000 episodes, whereas a frozen transferred memory does not. Inflating the memory to 100{,}000 rules keeps retrieval under 0.8\,ms, and the 1.59\,s miss cycle, dominated by the LLM call, runs off the control path behind a 2.6\,ms verified fallback.
\endgroup
\fi

\vspace{-2mm}
\section{\texorpdfstring{\rev{Limitations}}{Limitations}}
\label{sec:limitations}
\vspace{-1.5mm}

\begingroup\revcolor
\ifarxiv
\else
%%%%%%%%%%%%%%%%%%%%%%%%% camera-ready: compact version %%%%%%%%%%%%%%%%%%%%%%
\textbf{Abstraction gap and action interface.} The evaluation sits at the tactical decision layer: perception is assumed as clean factored variables, highD and rounD run in replay (reactive-replay variant in \appref{app:reactive_noise}), and occlusion, tracking, and map errors are absent; the five meta-actions are the standard interface of this layer and of all eight baselines, and continuous maneuvers would change only the scorer's output head. The claims are harness results at this layer, not a deployable AV stack, and the episode-level collision rates of 5.7--9.5\% in deliberately stressed traffic are not deployment grade; the supported claim is relative (49--56\% below the best learned baseline).
\textbf{Correlational dynamics.} The learned dynamics are prior-constrained but not causally identified; verification is conservative, so misattribution degrades toward rejecting actions rather than accepting unsafe ones, yet its quality depends on dynamics fidelity (constant-velocity propagation raises CR by 48--49\%; 3 of the 57 acquired rules were over-specific in audit, \appref{app:verification_residual}).
\textbf{Limited zero-shot cross-domain reusability.} Zero-shot cross-domain reusability of the compiled memory is limited: a memory compiled solely from highD/rounD begins at an 86.2\% call rate on highway-env, indicating that many target-domain encounters still require online acquisition. REINS is accordingly scoped to the self-evolution loop rather than assuming universal cross-domain memory transfer (\appref{app:transfer}).
\textbf{Scope.} The monotone miss-rate property assumes a stationary scene distribution, and the implementation is specialized to driving: only the variable schema and the action space must be re-authored for another domain (\appref{app:scope}).
\fi
\endgroup

\vspace{-2mm}
\section{Conclusion}
\vspace{-1.5mm}
\begingroup\revcolor We present REINS, a self-evolving agent harness that adapts the agent-harness paradigm from software engineering to real-time, safety-critical trajectory planning through three LLM--planner couplings: grounding, verification, and consolidation. Familiar scenes resolve from an indexed rule memory in $\mathcal{O}(\log|\mathcal{R}|)$ time, while memory misses trigger LLM-driven rule acquisition that grounds, verifies, and writes back new rules, closing the self-evolution loop without any LLM weight update. At the tactical decision layer, REINS attains 98\%+ compliance, nearly 50\% collision reduction, and 1.9--4.6\,ms latency, with gains that survive matched baselines, observation noise, reactive replay, and cross-domain memory transfer. Future work will couple the harness to perception uncertainty and continuous maneuvers, extend it to multi-agent coordination, and move the learned dynamics toward interventional identification with incremental updates.\endgroup

%%%%%%%%%%%%%%%%%%%%%%%%%%%%%%%%%%%%%%%%%%%%%%%%%%%%%%%%%%%%%%%%%%%%%%%%%%%%%%%
% ACKNOWLEDGMENTS
%%%%%%%%%%%%%%%%%%%%%%%%%%%%%%%%%%%%%%%%%%%%%%%%%%%%%%%%%%%%%%%%%%%%%%%%%%%%%%%
\begin{ack}
\revcolor This work was supported by the European Union's Horizon Europe research and innovation programme under Marie Sk\l{}odowska-Curie grant agreement No.~101126636 (DSTrain) and by the Technical Faculty of IT and Design at Aalborg University. This work was partly supported by the European Union under Grant~101123471 (EDGE-Skills project funded under the Digital Europe Programme). Views and opinions expressed are however those of the author(s) only and do not necessarily reflect those of the European Union or the European Commission. Neither the European Union nor the granting authority can be held responsible for them. The authors declare no competing interests.
\end{ack}

%%%%%%%%%%%%%%%%%%%%%%%%%%%%%%%%%%%%%%%%%%%%%%%%%%%%%%%%%%%%%%%%%%%%%%%%%%%%%%%
% REFERENCES
%%%%%%%%%%%%%%%%%%%%%%%%%%%%%%%%%%%%%%%%%%%%%%%%%%%%%%%%%%%%%%%%%%%%%%%%%%%%%%%
\bibliographystyle{plainnat}
\bibliography{example_paper}

%%%%%%%%%%%%%%%%%%%%%%%%%%%%%%%%%%%%%%%%%%%%%%%%%%%%%%%%%%%%%%%%%%%%%%%%%%%%%%%
% APPENDIX
%%%%%%%%%%%%%%%%%%%%%%%%%%%%%%%%%%%%%%%%%%%%%%%%%%%%%%%%%%%%%%%%%%%%%%%%%%%%%%%
\newpage
\appendix
%%%%%%%%%%%%%%%%%%%%%%%%%%%%%%%%%%%%%%%%%%%%%%%%%%%%%%%%%%%%%%%%%%%%%%%%%%%%%%%
% Appendix A: More Details about REINS
%%%%%%%%%%%%%%%%%%%%%%%%%%%%%%%%%%%%%%%%%%%%%%%%%%%%%%%%%%%%%%%%%%%%%%%%%%%%%%%
\section{More Details about REINS}
\label{app:clear_details}

\subsection{Variable Definitions}
\label{app:variables}
We define macro and micro variables shown in Table~\ref{tab:macro-micro}. The macro variables represent high-level semantic categories, while micro variables are the specific measurable quantities used in the world dynamics model and rule system.

The variables are organized into five functional groups:
\textbf{Environment} variables capture external conditions (weather, visibility, road surface, speed limits) that affect driving but are not directly controllable.
\textbf{Traffic} variables describe spatial and temporal relationships with surrounding vehicles, including headway distances, relative speeds, TTC, and lane density metrics.
\textbf{Ego} variables represent the controlled vehicle's kinematic state (position, velocity, acceleration, orientation).
\textbf{Action} defines the discrete action space: five high-level maneuvers executed by low-level controllers.
\textbf{Outcome} variables measure action consequences, including safety (collision risk), comfort (jerk-based), and task completion indicators.
This hierarchical organization enables structured rule retrieval via variable-group indexing.

The macro-group dependency set $\mathcal{E}_{\mathcal{H}}$ in Def.~\ref{def:variable_structure} encodes the admissible group-level edges:
\begin{equation}
\begin{aligned}
\mathcal{E}_{\mathcal{H}} = \{
&(\textsc{Env}, \textsc{Traffic}), (\textsc{Env}, \textsc{Ego}), (\textsc{Env}, \textsc{Outcome}), \\
&(\textsc{Traffic}, \textsc{Traffic}), (\textsc{Traffic}, \textsc{Ego}), (\textsc{Traffic}, \textsc{Outcome}), \\
&(\textsc{Ego}, \textsc{Traffic}), (\textsc{Ego}, \textsc{Ego}), (\textsc{Ego}, \textsc{Outcome}), \\
&(\textsc{Action}, \textsc{Ego}), (\textsc{Action}, \textsc{Outcome}) \}.
\end{aligned}
\label{eq:valid_edges}
\end{equation}
Environment factors influence Traffic, Ego, and Outcome; Traffic and Ego variables interact bidirectionally and feed into Outcome; Action affects only Ego state and Outcome.

\begin{table*}[!tbp]
\centering
\setlength{\tabcolsep}{2pt}
{
\fontsize{8.4pt}{10pt}\selectfont
\begin{tabular}{l|l|l}
\toprule
\textbf{Macro Variable} & \textbf{Micro Variable} & \textbf{Description} \\
\midrule
\multirow{6}{*}{\textbf{Environment}}
& weather\_type & Weather condition (clear/rain/fog/snow) \\
& visibility\_distance & Visibility range (m) \\
& road\_friction & Road surface friction coefficient \\
& road\_type & Scenario type (highway/merge/intersection/roundabout) \\
& time\_of\_day & Time period (day/night/dawn/dusk) \\
& speed\_limit & Posted speed limit (km/h) \\
\midrule
\multirow{22}{*}{\textbf{Traffic}}
& headway\_distance & Distance to lead vehicle (m) \\
& lead\_speed & Lead vehicle speed (m/s) \\
& lead\_rel\_speed & Relative speed to lead vehicle (m/s) \\
& lead\_ttc & Time-to-collision with lead vehicle (s) \\
& lead\_thw & Time headway to lead vehicle (s) \\
& rear\_distance & Distance to rear vehicle (m) \\
& rear\_speed & Rear vehicle speed (m/s) \\
& rear\_rel\_speed & Relative speed to rear vehicle (m/s) \\
& rear\_ttc & Time-to-collision with rear vehicle (s) \\
& left\_front\_distance & Distance to left-front vehicle (m) \\
& left\_front\_speed & Left-front vehicle speed (m/s) \\
& left\_rear\_distance & Distance to left-rear vehicle (m) \\
& left\_rear\_speed & Left-rear vehicle speed (m/s) \\
& right\_front\_distance & Distance to right-front vehicle (m) \\
& right\_front\_speed & Right-front vehicle speed (m/s) \\
& right\_rear\_distance & Distance to right-rear vehicle (m) \\
& right\_rear\_speed & Right-rear vehicle speed (m/s) \\
& lane\_density\_left & Left lane density (vehicles/100m) \\
& lane\_density\_right & Right lane density (vehicles/100m) \\
& available\_gap\_left & Available gap in left lane (m) \\
& available\_gap\_right & Available gap in right lane (m) \\
& cutin\_risk & Cut-in risk indicator (0--1) \\
\midrule
\multirow{10}{*}{\textbf{Ego}}
& ego\_x & Ego vehicle longitudinal position (m) \\
& ego\_y & Ego vehicle lateral position (m) \\
& ego\_vx & Ego vehicle longitudinal velocity (m/s) \\
& ego\_vy & Ego vehicle lateral velocity (m/s) \\
& ego\_speed & Ego vehicle total speed (m/s) \\
& ego\_acceleration & Ego vehicle acceleration (m/s$^2$) \\
& yaw\_rate & Ego vehicle yaw rate (rad/s) \\
& jerk & Ego vehicle jerk (m/s$^3$) \\
& lane\_id & Current lane identifier \\
& lateral\_offset & Lateral offset from lane center (m) \\
\midrule
\textbf{Action}
& action\_type & Discrete action (IDLE/FASTER/SLOWER/LANE\_LEFT/LANE\_RIGHT) \\
\midrule
\multirow{5}{*}{\textbf{Outcome}}
& risk\_score & Collision probability (0--1) \\
& comfort\_score & Comfort metric (jerk-based) \\
& energy\_consumption & Energy usage metric \\
& success\_flag & Goal completion indicator (boolean) \\
& violation\_flag & Rule violation indicator (boolean) \\
\bottomrule
\end{tabular}
}
\caption{Macro and micro variable definitions used in REINS.}
\label{tab:macro-micro}
\end{table*}

\subsection{Rule Templates}
\label{app:rules}
Seven rule templates are defined to capture different logical structures for driving knowledge:

\begin{itemize}[leftmargin=*,nosep]
    \item \texttt{\textbf{IF-THEN}}: Conditional implication for basic conditional relations.\\
    \textit{Example: IF TTC $<$ 2s THEN decelerate}

    \item \textbf{\texttt{AND-THEN}}: Conjunctive trigger requiring multiple conditions.\\
    \textit{Example: IF visibility $<$ 50m AND raining THEN reduce\_speed\_by\_30\%}

    \item \textbf{\texttt{THRESHOLD}}: Numeric constraint with calibrated boundaries.\\
    \textit{Example: IF speed $>$ speed\_limit THEN violation = true}

    \item \textbf{\texttt{PRIORITY}}: Decision hierarchy for conflicting actions.\\
    \textit{Example: yield $\succ$ accelerate (yield takes priority)}

    \item \textbf{\texttt{MULTI-ACTION}}: Compound actions under one condition.\\
    \textit{Example: IF collision\_risk = high THEN brake AND steer\_away}

    \item \textbf{\texttt{WEIGHTED-ACTION}}: Weighted action selection.\\
    \textit{Example: IF crash\_risk THEN \{brake: 0.7, steer: 0.3\}}

    \item \textbf{\texttt{TEMPORAL-CHAIN}}: Temporal constraints between triggers.\\
    \textit{Example: IF obstacle\_detected THEN brake within 1.5s}
\end{itemize}

\subsection{Rule Conflict Resolution}
\label{app:conflict}
When multiple rules are activated simultaneously, conflicts may arise if they recommend different actions. REINS resolves conflicts through a three-level arbitration mechanism:

\textbf{Priority-based ordering.} Rules are assigned priority levels based on their functional category: Safety $>$ Compliance $>$ Navigation $>$ Efficiency. When rules from different categories conflict, the higher-priority rule takes precedence. For example, if an Efficiency rule (consider lane change) and a Safety rule (maintain gap) both fire, the Safety rule overrides.

\textbf{Confidence-weighted aggregation.} For rules within the same priority level, REINS aggregates their scores using the noisy-OR formula (Eq.~\eqref{eq:prior}). This allows multiple supporting rules to reinforce an action while remaining robust to individual noisy predictions. The action with the highest aggregated probability is selected.

\textbf{Dynamics-based verification.} When the aggregated action prior is below a confidence threshold ($\tau = 0.7$), REINS falls back to dynamics-based verification. The top-$M$ candidate actions are simulated via forward rollout using the world dynamics model, and the action yielding the safest predicted trajectory is selected.

This hierarchical approach ensures that safety-critical rules are never overridden by efficiency considerations, while still allowing flexible action selection in non-critical situations.

\subsection{Method Details}
\label{app:method_details}

This subsection contains the full equations, architectural details, and explanatory prose deferred from the methodology to keep the main body concise. The structure mirrors Section~\ref{sec:offline}--\ref{sec:online}.

\subsubsection{World Dynamics Learning Details}
\label{app:world_dynamics_details}

We learn the dependency graph $\hat{\mathcal{G}} = (\mathcal{X}, \hat{\mathcal{E}}, \hat{w})$ in three steps, then train a per-variable dynamics function on the resulting parent sets.

\textbf{Step 1: Edge Prior Estimation.} We query the LLM with variable dependency prompts (Appendix~\ref{app:prompts}) to obtain an edge confidence $\pi_{ij} \in [0,1]$ for each candidate edge $X_j \to X_i$. This is combined with the macro-group constraints (Eq.~\ref{eq:valid_edges}) to form a unified prior:
\begin{equation}
\tilde{\pi}_{ij} = \alpha\,\pi_{ij} + (1-\alpha)\,\mathbb{I}\{(h(X_j), h(X_i)) \in \mathcal{E}_{\mathcal{H}}\},
\end{equation}
where $\alpha \in [0,1]$ balances LLM confidence against structural constraints. Edges that violate the macro-group constraints carry indicator $0$ and are reduced to $\alpha\,\pi_{ij}$, while semantically aligned edges are pushed toward $1$.

\textbf{Step 2: Prior-Guided Regression.} For each target variable $X_i$, we predict its value from the lagged values of other variables via ridge regression~\citep{hoerl1970ridge} with prior-adaptive penalties $\lambda_{ij} = \lambda_0(1 - \tilde{\pi}_{ij})^\rho$:
\begin{equation}
\hat{\mathbf{w}}_i = \arg\min_{\mathbf{w}} \sum_{t=L}^N \Big(X_i^t - \textstyle\sum_{j\neq i}\sum_{\ell=1}^L w_{j\ell} X_j^{t-\ell}\Big)^2 + \sum_{j\neq i}\lambda_{ij}\|\mathbf{w}_j\|_2^2,
\end{equation}
where $w_{j\ell}$ is the regression coefficient for variable $X_j$ at lag $\ell \in \{1, \ldots, L\}$, $\mathbf{w}_j = (w_{j1}, \ldots, w_{jL})^\top$ groups all lags for $X_j$, $\lambda_0$ is the base penalty, and $\rho > 0$ controls penalty sharpness. High-prior edges ($\tilde{\pi}_{ij} \to 1$) receive weak regularization and survive; low-prior edges are shrunk toward zero.

\textbf{Step 3: Graph Construction.} The parent set $\mathrm{Pa}(X_i)$ retains the top-$K$ variables ranked by aggregated coefficient magnitude $\sum_{\ell=1}^L |\hat{w}_{j\ell}|$. The learned edge set and weights are
\begin{equation}
\hat{\mathcal{E}} = \{(X_j, X_i) \mid X_j \in \mathrm{Pa}(X_i)\}, \quad \hat{w}(X_j, X_i) = \textstyle\sum_{\ell=1}^L |\hat{w}_{j\ell}|.
\label{eq:graph_construct}
\end{equation}

\textbf{Dynamics Function Loss.} Given $\hat{\mathcal{G}}$, each $f_i$ is implemented as a multi-head Transformer~\citep{vaswani2017attention} and trained by mean squared error over its parent trajectory:
\begin{equation}
\mathcal{L}_{\text{dyn}}^{(i)} = \frac{1}{N} \sum_{t} \big(X_i^{t+1} - f_i(\mathrm{Pa}(X_i)^{t-L+1:t})\big)^2.
\label{eq:dyn_loss}
\end{equation}
Optimizer, batch size, and convergence curves are reported in Appendix~\ref{app:training_loss}.

\subsubsection{Rule Compilation Details}
\label{app:rule_compilation_details}

\textbf{Rule Structure Generation.} For each dependency edge $(X_j, X_i) \in \hat{\mathcal{E}}$, the LLM generates a rule structure
\begin{equation}
r = (T, \mathcal{P}_r, a_r, c_r),
\label{eq:rule_gen}
\end{equation}
where $T$ is the rule template (Def.~\ref{def:rule_template}), $\mathcal{P}_r$ is a set of guard predicates with symbolic thresholds, $a_r \in \mathcal{A}$ is the recommended action (Def.~\ref{def:action}), and $c_r$ is the confidence derived from edge weight $\hat{w}(X_j, X_i)$. Prompt details are in Appendix~\ref{app:prompts}.

\textbf{Threshold Calibration.} Each predicate $p \in \mathcal{P}_r$ (Def.~\ref{def:predicate}) requires a concrete threshold $\tau_i$, calibrated from $\mathcal{D}$ in three forms~\citep{luo2024graph}:
\begin{equation}
\tau_i = \begin{cases} \mu_i + \zeta\sigma_i & \text{(z-score)},\\ Q_i(\zeta) & \text{(quantile)},\\ \tau_0 & \text{(absolute)}, \end{cases}
\end{equation}
where $\mu_i, \sigma_i$ are empirical mean and standard deviation of $X_i$, $Q_i(\zeta)$ is the $\zeta$-quantile, and $\tau_0$ is a fixed constant (e.g., speed limits). The comparison operator $\prec$ is selected to maximize safety separability between the two sides of the threshold under safety outcome $y \in \{0,1\}$ (Section~\ref{sec:rule_compilation}).

\begin{figure}[htbp]
    \centering
    \includegraphics[width=0.6\linewidth]{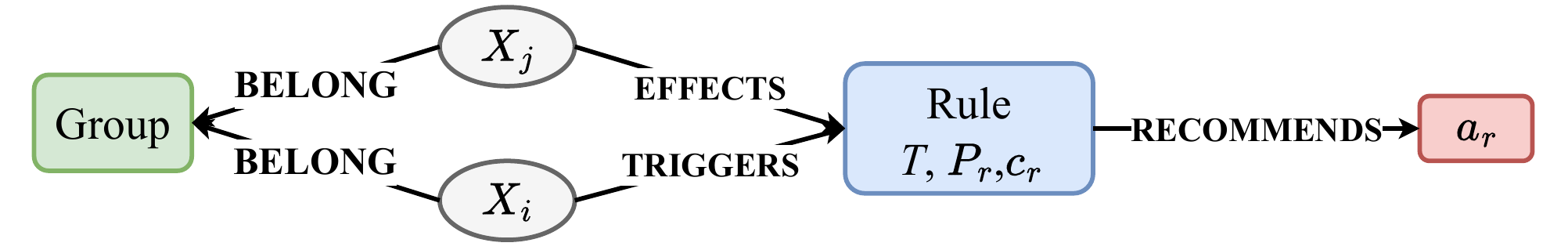}
    \caption{Rule database schema.}
    \label{fig:rule_database_schema}
\end{figure}

\textbf{Graph Database Storage.} Compiled rules are stored in a Neo4j graph database~\citep{webber2012programmatic} with a variable-indexed schema (Figure~\ref{fig:rule_database_schema}). Each rule node stores the complete tuple $(T, \mathcal{P}_r, a_r, c_r)$, where $T$ is the template type, $\mathcal{P}_r$ contains calibrated predicates with concrete thresholds $\tau$ and operators $\prec$, $a_r$ is the recommended action, and $c_r$ is the confidence. Rule nodes connect to variable nodes via \texttt{TRIGGERS} (input variables) and \texttt{AFFECTS} (output variables) edges. This enables $O(\log|\mathcal{R}|)$ retrieval via parameterized Cypher queries on variables active in the current state. Database statistics are provided in Appendix~\ref{app:rule_database_statistics}.

\subsubsection{Rule-Guided Planning Details}
\label{app:planning_details}

\textbf{Scene-Aware Rule Retrieval.} Given $s_t \in \mathcal{S}$, we identify active variables $\{X_i\}$ and retrieve context-relevant rules via parameterized Cypher queries~\citep{francis2018cypher} generated by the LLM (read-only and schema-constrained; Appendix~\ref{app:prompts}). The retrieval map
\begin{equation}
G: (\{X_i\}, K) \longmapsto \mathcal{R}_t \subseteq \mathcal{R}
\label{eq:cypher_gen}
\end{equation}
returns up to $K$ rules ranked by confidence $c_r$. A rule $r$ is retrieved if its trigger variables $\{X_j \mid (X_j, r) \in \texttt{TRIGGERS}\}$ intersect with the active set $\{X_i\}$ and the dependency direction satisfies the macro-group constraints $\mathcal{E}_{\mathcal{H}}$.

\textbf{Action Scorer Architecture.} The Transformer scorer~\citep{liang2022trajformer} takes state encoding $\mathbf{x}_t$ and rule embedding $\mathbf{u}_r$ and emits relevance and action heads:
\begin{equation}
\begin{aligned}
\mathbf{h}_{t,r} &= \text{TransformerEnc}([\mathbf{x}_t; \mathbf{u}_r]), \\
S_{\text{rel}}(r, s_t) &= \sigma(\text{Dec}_{\text{rel}}(\mathbf{h}_{t,r})) \in [0,1], \\
S_{\text{act}}(r, s_t, a) &= \mathrm{softmax}(\text{Dec}_{\text{act}}(\mathbf{h}_{t,r}))_a,
\end{aligned}
\label{eq:rule_score}
\end{equation}
where $S_{\text{rel}}$ predicts whether $r$ should influence the current decision, and $S_{\text{act}}$ predicts the action preference given $r$. The combined rule-action score multiplies relevance, action preference, and rule confidence:
\begin{equation}
f(r, a \mid s_t) = S_{\text{rel}}(r, s_t) \cdot S_{\text{act}}(r, s_t, a) \cdot c_r.
\label{eq:rule_action_score}
\end{equation}

\textbf{Scorer Training.} The scorer is trained on expert trajectories $\{(s_t, a^*_t)\}$ with a joint relevance + action objective:
\begin{equation}
\mathcal{L} = \sum_{t,r}\big[\mathrm{BCE}(\mathrm{Exec}(r, s_t), S_{\text{rel}}(r, s_t)) - \log S_{\text{act}}(r, s_t, a^*_t)\big],
\label{eq:scorer_loss}
\end{equation}
where $\mathrm{Exec}(r, s_t) \in \{0,1\}$ is the ground-truth rule activation (Eq.~\ref{eq:rule_exec}) and $a^*_t$ is the expert action.

\textbf{Action Aggregation.} The full noisy-OR action distribution corresponding to Eq.~\ref{eq:prior} is
\begin{equation}
\begin{aligned}
P(a \mid s_t) &= \frac{1 - \prod_{r \in \mathcal{R}_t(a)}\big(1 - f(r, a \mid s_t)\big)}{Z},\\
a^*_{\text{rule}} &= \arg\max_{a \in \mathcal{A}} P(a \mid s_t),
\end{aligned}
\end{equation}
where $\mathcal{R}_t(a) \subseteq \mathcal{R}_t$ denotes the rules with recommended action $a_r = a$ and $Z$ normalizes over $\mathcal{A}$. The formulation favors actions supported by multiple high-confidence rules.

\textbf{Trajectory Verification.} For the top-$M$ actions ranked by $P(a \mid s_t)$, we predict $H$-step future states:
\begin{equation}
\hat{s}_{t+h+1} = \Phi(\hat{s}_{t+h}, a; \hat{\mathcal{G}}, \hat{\mathcal{F}}), \quad h = 0, \ldots, H-1,
\label{eq:rollout}
\end{equation}
where $\Phi$ applies each $f_i \in \hat{\mathcal{F}}$ to predict $X_i$ from $\mathrm{Pa}(X_i)$ under $\hat{\mathcal{G}}$. Each rollout produces a candidate trajectory $\hat{\xi}(a) = (\hat{s}_{t+1}, \ldots, \hat{s}_{t+H})$, consistent with Def.~\ref{def:dynamics}. The planner selects $a^\star_t = \arg\max_{a \in \mathcal{A}_M} J(\hat{\xi}(a))$, where $J: \mathcal{S}^H \to \mathbb{R}$ penalizes collisions, rule violations, and control effort (Appendix~\ref{app:metrics}); the first control of the selected trajectory is executed in receding-horizon~\citep{zhang2021world}.

\subsubsection{Online Rule Acquisition Details}
\label{app:acquisition_details}

\textbf{LLM-Driven Rule Proposal.} On a miss, the harness queries $\mathcal{M}$ with the current state and active variables:
\begin{equation}
\tilde{r} = (\tilde{T}, \tilde{\mathcal{P}}, \tilde{a}, \tilde{c}) \;\sim\; \mathcal{M}(s_t),
\label{eq:llm_proposal}
\end{equation}
where $\tilde{r}$ conforms to the rule template (Def.~\ref{def:rule_template}) and uses only variables in $\mathcal{X}$ and operators from the predicate grammar (Def.~\ref{def:predicate}). Prompt details are in Appendix~\ref{app:prompts}.

\textbf{Safe-Action Selection.} When online rule acquisition rejects $\tilde{r}$ and $\mathcal{R}_t = \emptyset$, the harness raises a human-takeover alert and executes the action that maximizes minimum lookahead time-to-collision under $\hat{\mathcal{F}}$:
\begin{equation}
a^{\mathrm{safe}}_t = \arg\max_{a \in \mathcal{A}}\;\min_{h \in [1, H]}\mathrm{TTC}\!\left(\hat{s}_{t+h}(a)\right).
\label{eq:safe_action}
\end{equation}

\textbf{Memory-Update Collision Branch.} Eq.~\ref{eq:memory_update}'s collision branch averages confidences instead of creating a duplicate node, so $|\mathcal{R}|$ counts distinct rule structures rather than LLM-noise replicas. The hash $\kappa$ is computed over the structural triple $(\tilde{T}, \tilde{\mathcal{P}}, \tilde{a})$ but excludes $\tilde{c}$, so semantically identical proposals collide regardless of confidence noise. Collisions are common during early deployment when $\mathcal{R}$ is small, then taper off as coverage saturates.

\textbf{Self-Evolution Properties.} Eq.~\ref{eq:memory_update} structurally enforces $|\mathcal{R}_{t+1}| \geq |\mathcal{R}_t|$ (memory monotonicity) and $\theta_{\mathcal{M}} \equiv \theta_{\mathcal{M}}^{(0)}$ (no LLM parameter update). Under stationarity of the scene distribution, the per-state miss rate is non-increasing as $\mathcal{R}$ accumulates coverage, since any state $s$ that triggered a miss and produced an accepted $\tilde{r}$ is now matched by a rule keyed on the variables active at $s$. Since $\mathcal{M}$ is invoked only on misses, its per-scene invocation rate decays monotonically with $\Pr[\mathrm{Miss}]$. We validate this behavior empirically in Section~\ref{sec:self_evolution} and Appendix~\ref{app:self_evolution_extended}.

\subsection{Prompts}
\label{app:prompts}

We provide three prompt templates used in REINS's offline compilation pipeline. Figure~\ref{fig:prompt-dependency} shows the \textbf{Variable Dependency Prompt}, which instructs the LLM to infer dependencies between driving variables based on domain knowledge; the output serves as semantic priors for structure learning in the world dynamics model. Figure~\ref{fig:prompt-ruleextraction} presents the \textbf{Rule Extraction Prompt}, which converts dependency edges into executable driving rules with parameterized guards and discrete actions; these rules are stored in the Neo4j graph database. Figure~\ref{fig:prompt-cypherretrieval} displays the \textbf{Parameterized Cypher Query Prompt}, which generates parameterized queries for context-aware rule retrieval during planning. All prompts enforce read-only constraints and structured output formats to ensure safety and parseability.

\begin{figure*}[htp]
\centering
\begin{tcolorbox}[
  colback=gray!8, colframe=black!80,
  boxrule=0.5pt, arc=3mm,
  left=8pt, right=8pt, top=6pt, bottom=6pt,
  title=Variable Dependency Prompt,
  colbacktitle=gray!40, coltitle=black, fonttitle=\bfseries
]
[\textbf{TASK}]\\
You are \textbf{an expert in autonomous driving systems and variable dependency reasoning}.\\
Your task is to identify a set of \textbf{plausible dependency relationships} among driving-related variables observed in an autonomous vehicle simulation environment. \\
Each relationship should describe how one variable may influence another under realistic driving conditions.\\
These dependency relations will be used to construct a \textbf{dependency graph} for downstream decision-making and trajectory planning.

Each dependency relationship must be:\\
$\bullet$ \textbf{Physically plausible}: reflecting physical laws, driving behavior, or traffic context;\\
$\bullet$ \textbf{Directional}: clearly indicating the influencing variable and the affected variable;\\
$\bullet$ \textbf{Actionable}: relevant to planning, safety, efficiency, or comfort;\\
$\bullet$ \textbf{Concise}: expressed as a simple directed pair without explanation.

[\textbf{INPUT}]\\
You are given a set of driving-related variables extracted from a simulation environment:\\
$\bullet$ Variable list: \texttt{\{variable\_list\}}

[\textbf{OUTPUT}]\\
Please output a list of dependency relationships in the following strict format. \\
Each relationship must be written as:
\[
\texttt{SOURCE} \rightarrow \texttt{TARGET}
\]
Return all relationships enclosed in triple quotes, and do not include any explanations or additional text.\\
\verb|'''|\\
SOURCE $\rightarrow$ TARGET\\
SOURCE $\rightarrow$ TARGET\\
SOURCE $\rightarrow$ TARGET\\
\verb|'''|

[\textbf{EXAMPLE}]\\
\verb|'''|\\
ego\_speed $\rightarrow$ energy\_consumption\\
road\_friction $\rightarrow$ risk\_score\\
headway\_distance $\rightarrow$ cutin\_risk\\
\verb|'''|

Make sure that your output strictly follows this format. Any relationship that does not adhere to the \texttt{SOURCE} $\rightarrow$ \texttt{TARGET} structure should be removed.
\end{tcolorbox}
\caption{Variable dependency prompt. Given a list of driving-related variables, the LLM infers directed dependency relationships as structural priors for the world dynamics model.}
\label{fig:prompt-dependency}
\end{figure*}

\begin{figure*}[htp]
\centering
\begin{tcolorbox}[
  colback=gray!8, colframe=black!80,
  boxrule=0.5pt, arc=3mm,
  left=8pt, right=8pt, top=6pt, bottom=6pt,
  title=Rule Extraction Prompt,
  colbacktitle=gray!40, coltitle=black, fonttitle=\bfseries
]
\footnotesize
[\textbf{TASK}]\\
You are an expert in autonomous driving and rule-based reasoning.\\
Given dependency edges from a learned dependency graph, extract executable driving rules that can be stored in a graph database and used for trajectory planning.

Each rule must be:\\
$\bullet$ \textbf{Grounded}: based on the learned dependency structure;\\
$\bullet$ \textbf{Executable}: containing guards with parameterizable thresholds;\\
$\bullet$ \textbf{Actionable}: mapping to discrete actions (LANE\_LEFT, LANE\_RIGHT, IDLE, FASTER, SLOWER).

[\textbf{INPUT}]\\
You are given dependency edges with learned weights from the dependency graph:\\
$\bullet$ Dependency edges: \texttt{\{dependency\_edges\}}\\
$\bullet$ Variable descriptions: \texttt{\{variable\_descriptions\}}

[\textbf{OUTPUT FORMAT}]\\
Return a JSON array of rule objects. Each object must include:
\[
\texttt{\{"rule\_index", "rule\_type", "variables", "params", "guards", "action", "confidence"\}}
\]

\begin{verbatim}
{
 "rule_index": "R_xxx",
  "rule_type": "physical|behavioral|scene",
  "variables": ["cause_var", "effect_var"],
  "params": {"threshold_form": "z|q|abs", "k": <float>},
  "guards": ["cause_var <op> threshold"],
  "action": "SLOWER|FASTER|LANE_LEFT|LANE_RIGHT|IDLE",
  "confidence": <float in [0,1]>
}
\end{verbatim}

[\textbf{EXAMPLE}]\\
\textit{Input}: headway\_distance $\rightarrow$ risk\_score (weight=0.85)\\
\textit{Output}:
\begin{verbatim}
{"rule_index": "R_001", "rule_type": "behavioral",
 "variables": ["headway_distance", "risk_score"],
 "params": {"threshold_form": "z", "k": -1.5},
 "guards": ["headway_distance < threshold"],
 "action": "SLOWER", "confidence": 0.85}
\end{verbatim}

Make sure each rule follows the format strictly and maps to one of the five discrete actions.
\end{tcolorbox}
\caption{Rule extraction prompt from dependency structure.}
\label{fig:prompt-ruleextraction}
\end{figure*}

\begin{figure*}[htp]
\centering
\begin{tcolorbox}[
  colback=gray!8, colframe=black!80,
  boxrule=0.5pt, arc=3mm,
  left=8pt, right=8pt, top=6pt, bottom=6pt,
  title=Parameterized Cypher Query Prompt,
  colbacktitle=gray!40, coltitle=black, fonttitle=\bfseries
]
\footnotesize
[\textbf{TASK}]\\
You are an expert in Neo4j graph databases and autonomous driving systems.\\
Generate a parameterized, \textbf{read-only} Cypher query to retrieve relevant driving rules based on the current scene context.

[\textbf{INPUT}]\\
You are given the following information about the current driving scene:\\
$\bullet$ Scene summary: \texttt{\{scene\_summary\}}\\
$\bullet$ Active cause variables: \texttt{\{active\_causes\}}\\
$\bullet$ Target variable groups: \texttt{\{target\_groups\}} (e.g., Outcome, Ego, Action)\\
$\bullet$ Maximum results: \texttt{\{k\}}

[\textbf{GRAPH SCHEMA}]\\
The rule database contains the following structure:\\
$\bullet$ Nodes: \texttt{(Rule)}, \texttt{(Variable)}, \texttt{(Category)}, \texttt{(Group)}, \texttt{(Action)}\\
$\bullet$ Relations: \texttt{[:TRIGGERS]}, \texttt{[:CAUSES]}, \texttt{[:BELONG\_TO]}

[\textbf{CONSTRAINTS}]\\
Your query must satisfy the following safety requirements:\\
$\bullet$ Query MUST start with: \texttt{MATCH}, \texttt{WITH}, \texttt{RETURN}, \texttt{CALL}, or \texttt{UNWIND}\\
$\bullet$ NO write operations: \texttt{CREATE}, \texttt{MERGE}, \texttt{SET}, \texttt{DELETE}, \texttt{DETACH}, \texttt{LOAD}, \texttt{DROP}, \texttt{ALTER}\\
$\bullet$ Use parameter placeholders: \texttt{\$active\_causes}, \texttt{\$target\_groups}, \texttt{\$k}

[\textbf{OUTPUT}]\\
Return only the Cypher query string enclosed in triple quotes. Do not include any explanations.\\
\begin{verbatim}
'''
<Cypher query here>
'''
\end{verbatim}

[\textbf{EXAMPLE}]\\
\begin{verbatim}
'''
MATCH (r:Rule)-[:TRIGGERS]->(c:Variable)-[:CAUSES*0..1]->(e:Variable)
WHERE c.name IN $active_causes AND e.group IN $target_groups
RETURN r.rule_id, r.text, collect(e.name) AS effects,
       r.type, r.confidence
ORDER BY r.confidence DESC LIMIT $k
'''
\end{verbatim}

Ensure the query is syntactically correct and adheres to all constraints.
\end{tcolorbox}
\caption{Parameterized Cypher query prompt.}
\label{fig:prompt-cypherretrieval}
\end{figure*}

%%%%%%%%%%%%%%%%%%%%%%%%%%%%%%%%%%%%%%%%%%%%%%%%%%%%%%%%%%%%%%%%%%%%%%%%%%%%%%%
% Appendix B: Implementation Details
%%%%%%%%%%%%%%%%%%%%%%%%%%%%%%%%%%%%%%%%%%%%%%%%%%%%%%%%%%%%%%%%%%%%%%%%%%%%%%%
\section{Implementation Details}
\label{app:implementation}

\subsection{Experiment Settings}
\label{app:exp_settings}

Table~\ref{tab:exp_settings} summarizes the key hyperparameters and training configurations used in our experiments. The settings are organized into six categories:
\textbf{Simulation Environment} specifies the \texttt{highway-env} simulator settings, including four driving scenarios with varying complexity and a discrete action space of five high-level maneuvers.
\textbf{Real-world Datasets} describes the highD (highway) and rounD (roundabout) drone datasets used for real-world evaluation, including recording statistics and sampling rates.
\textbf{World Dynamics Model} details the structure learning parameters (LLM prior weight, regularization) and neural network architecture (Transformer layers, attention heads).
\textbf{Rule Compilation} covers the offline LLM-based rule generation process, including the backbone model, sampling strategy, and threshold calibration.
\textbf{Training \& Evaluation} specifies the data splits, random seeds, and hardware configuration.
\textbf{Planning \& Deployment} reports inference-time parameters and latency measurements demonstrating real-time capability.
All hyperparameter configurations follow standard practices unless otherwise stated.

\begin{table}[htbp]
\centering
\footnotesize
\setlength{\tabcolsep}{1.5pt}
\begin{tabular}{l|l}
\toprule
\textbf{Category} & \textbf{Setting} \\
\midrule
\multicolumn{2}{c}{\textit{Simulation Environment}} \\
\midrule
Simulator & highway-env v1.8.2 \\
Scenarios & Highway, Merge, Roundabout, \\
 & Intersection \\
Episode length & 100 steps (20s at 5\,Hz) \\
Observation & 15 nearest vehicles \\
Action space & \{IDLE, FASTER, SLOWER, \\
 & LANE\_LEFT, LANE\_RIGHT\} \\
Weather conditions & Clear, Rain, Fog, Snow \\
\midrule
\multicolumn{2}{c}{\textit{Real-world Datasets}} \\
\midrule
highD source & German Autobahn (drone) \\
highD recordings & 01, 02, 03, 04 \\
highD vehicles & 4,237 (578 lane-change events) \\
rounD source & German roundabouts \\
rounD recordings & 00, 01, 02 \\
rounD vehicles & 2,156 (423 yield/merge events) \\
Sampling rate & 25\,Hz \\
\midrule
\multicolumn{2}{c}{\textit{World Dynamics Model}} \\
\midrule
Structure learning & Ridge regression + LLM prior \\
LLM prior weight $\alpha$ & 0.6 \\
Penalty exponent $\rho$ & 2.0 \\
Base regularization $\lambda_0$ & 0.1 \\
Lag length $L$ & 6 steps \\
Top-$K$ parents & 5 \\
Transformer layers & 2 \\
Attention heads & 4 \\
Optimizer & Adam \\
Learning rate & $2 \times 10^{-4}$ \\
Batch size & 64 \\
Training episodes & 50k \\
\midrule
\multicolumn{2}{c}{\textit{Rule Compilation (Offline)}} \\
\midrule
LLM backbone & GPT-4o (2024-08-06) \\
LLM temperature & 0.0 (deterministic) \\
Max output tokens & 4096 \\
Prompt tokens (avg.) & $\sim$1.2k per query \\
Total LLM calls & $\sim$50 (one-time offline) \\
Rule templates & 7 types \\
Neo4j version & 5.12 \\
Threshold calibration & Gaussian ($\mu+\zeta\sigma$), $\zeta$=1.5 \\
Confidence $\tau$ & 0.7 \\
\midrule
\multicolumn{2}{c}{\textit{Training \& Evaluation}} \\
\midrule
Training data & 50k episodes \\
Test episodes & 1000 per scenario \\
Random seeds & 5 (0, 42, 123, 456, 789) \\
Training hardware & 4$\times$ A100 (40GB) \\
Training time & $\sim$8 hours \\
\midrule
\multicolumn{2}{c}{\textit{Planning \& Deployment}} \\
\midrule
Retrieval count $k$ & 10 \\
Planning horizon $H$ & 5 steps \\
Rollout discount $\gamma$ & 0.99 \\
Safety weight $\lambda_{\text{safe}}$ & 1.0 \\
Comfort weight $\lambda_{\text{comf}}$ & 0.1 \\
Progress weight $\lambda_{\text{prog}}$ & 0.5 \\
Inference hardware & RTX 3090 or above \\
GPU memory & $\sim$4\,GB \\
Neo4j query latency & $<$1\,ms \\
Planning latency & 1.9--4.6\,ms per decision \\
Real-time capability & 200--500\,Hz \\
\bottomrule
\end{tabular}
\caption{Experiment settings and hyperparameters for REINS.}
\label{tab:exp_settings}
\end{table}

\subsection{Evaluation Metrics}
\label{app:metrics}

\textbf{Rule Compliance (Compl.)} measures the fraction of timesteps where all active rules are satisfied:
\begin{equation}
    \text{Compl.} = \frac{1}{T}\sum_{t=1}^{T}\prod_{k=1}^{K}\bigl(1-v_{k,t}\bigr),
\end{equation}
where $v_{k,t} \in \{0,1\}$ indicates whether rule $k$ is violated at time $t$.

\textbf{Collision Rate (CR)} measures the fraction of episodes in which at least one collision occurs:
\begin{equation}
    \text{CR} = \frac{1}{N}\sum_{i=1}^N \mathbb{I}(\text{collision in episode } i),
\end{equation}
where $N$ is the number of test episodes. A lower CR indicates safer planning.

\textbf{Risk-aware Efficiency (RE)} balances task performance against safety:
\begin{align}
\text{RE} &= R_{\text{task}}(\tau) - \lambda_{\text{risk}} R_{\text{risk}}(\tau), \\
R_{\text{risk}}(\tau) &= \sum_{t=1}^{T}\max(0, \delta_{\text{ttc}} - \text{TTC}_t),
\end{align}
where $R_{\text{task}}$ is the task reward, $R_{\text{risk}}$ penalizes low TTC states, $\lambda_{\text{risk}}=0.5$ is the risk penalty weight, and $\delta_{\text{ttc}}=3.0$s is the TTC threshold.

\textbf{Planning Efficiency (PE)} measures average computation time per decision:
\begin{equation}
    \text{PE} = \frac{1}{T}\sum_{t=1}^{T}t_{\text{plan}},
\end{equation}
where $t_{\text{plan}}$ (ms) is the wall-clock time for action selection.

\textbf{Trajectory Objective $J(\hat{\xi})$.} The trajectory-level objective used in dynamics-based verification is defined as:
\begin{equation}
\begin{aligned}
        J(\hat{\xi}) & = \sum_{h=1}^{H} \gamma^{h-1} \Big[ -\lambda_{\text{safe}} \cdot \mathbb{I}(\text{TTC}_{t+h} < \tau_{\text{ttc}}) \\
        & + \lambda_{\text{prog}} \cdot \Delta d_{t+h} - \lambda_{\text{comf}} \cdot |j_{t+h}| \Big],
\end{aligned}
\end{equation}
where $\text{TTC}_{t+h}$ is the predicted time-to-collision, $\Delta d_{t+h}$ is the progress toward the goal, $j_{t+h}$ is the jerk, and $\tau_{\text{ttc}} = 2.0$s is the safety threshold. The weights $\lambda_{\text{safe}}=1.0$, $\lambda_{\text{prog}}=0.5$, $\lambda_{\text{comf}}=0.1$ prioritize safety over progress and comfort.

\textbf{Real-world Evaluation Protocol.} We use closed-loop replay: one recorded vehicle is controlled by the planner while others follow recorded trajectories. Collisions are detected when bounding boxes overlap or the ego deviates beyond lane boundaries. \rev{In the simulated scenarios the surrounding vehicles instead react to the ego (IDM with MOBIL); a reactive-replay variant of highD is reported in Appendix~\ref{app:reactive_noise}.}

\subsection{Baseline Descriptions}
\label{app:baselines}

We select baselines from three categories to comprehensively evaluate REINS: (i) \emph{classical planners} (MPC, MCTS) that represent traditional optimization and search-based approaches without rule guidance; (ii) \emph{LLM-based planners} (DriveSpeak, PDDL, LLaDA, DriReg) that leverage language models for driving decisions but differ in how they integrate linguistic knowledge; and (iii) \emph{hybrid methods} (SymAgent, Plan-R1) that combine learning with symbolic reasoning. This selection covers the spectrum from rule-free to rule-guided planning, enabling analysis of how different rule representation and retrieval strategies affect safety and efficiency.

\begin{itemize}[leftmargin=*,nosep]
\item \textbf{MPC}~\citep{falcone2007predictive}: Classical optimization-based planner solving finite-horizon control under kinematic and safety constraints. We use a 5-step horizon with quadratic cost on acceleration and jerk.

\item \textbf{MCTS}~\citep{hubmann2016decentralized}: Sampling-based planning via tree search and rollout evaluation. We use 1000 simulations per decision with UCB exploration.

\item \textbf{DriveSpeak}~\citep{cui2024drive}: LLM-based method mapping natural-language scene descriptions directly to driving actions. Uses GPT-4o with scene prompts at each timestep.

\item \rev{\textbf{PDDL}~\citep{guan2023leveraging,liu2023llmp}: the LLM writes the domain's symbolic preconditions and effects once offline; at runtime a classical STRIPS solver (Pyperplan~\citep{alkhazraji2020pyperplan}) is invoked in process at each decision step with no LLM call. Its runtime profile is in Appendix~\ref{app:pddl_profile}.}

\item \textbf{LLaDA}~\citep{li2024driving}: LLM interpretation of region-specific traffic rules from driver handbooks. Uses pre-extracted rule sets with LLM-based rule matching.

\item \textbf{DriReg}~\citep{cai2024driving}: RAG-based traffic regulation retrieval with LLM reasoning for action legality. Retrieves top-5 regulations per query using embedding similarity.

\item \textbf{SymAgent}~\citep{liu2025symagent}: Neural-symbolic agent integrating LLMs with structured knowledge graphs. Uses the same Neo4j backend as REINS for fair comparison.

\item \textbf{Plan-R1}~\citep{tang2025plan}: Two-stage learning-based planner with rule-shaped rewards and reinforcement learning. Trained on the same 50k episodes as REINS's world model.
\end{itemize}

All baselines use identical observation spaces, action spaces, episodes, and evaluation protocols. For LLM-based methods, we use GPT-4o (2024-08-06) with temperature 0 for deterministic outputs. Hyperparameters for each baseline follow their original papers or are tuned on a held-out validation set. \rev{The conditions that are shared with REINS and those that each rule-based baseline keeps by design, together with a matched re-run and a common-reference compliance audit, are documented in Appendix~\ref{app:fairness}.}

%%%%%%%%%%%%%%%%%%%%%%%%%%%%%%%%%%%%%%%%%%%%%%%%%%%%%%%%%%%%%%%%%%%%%%%%%%%%%%%
% Appendix C: Additional Results
%%%%%%%%%%%%%%%%%%%%%%%%%%%%%%%%%%%%%%%%%%%%%%%%%%%%%%%%%%%%%%%%%%%%%%%%%%%%%%%
\section{Additional Results}
\label{app:additional}

\begin{figure*}[htbp]
    \centering
    \includegraphics[width=1\textwidth]{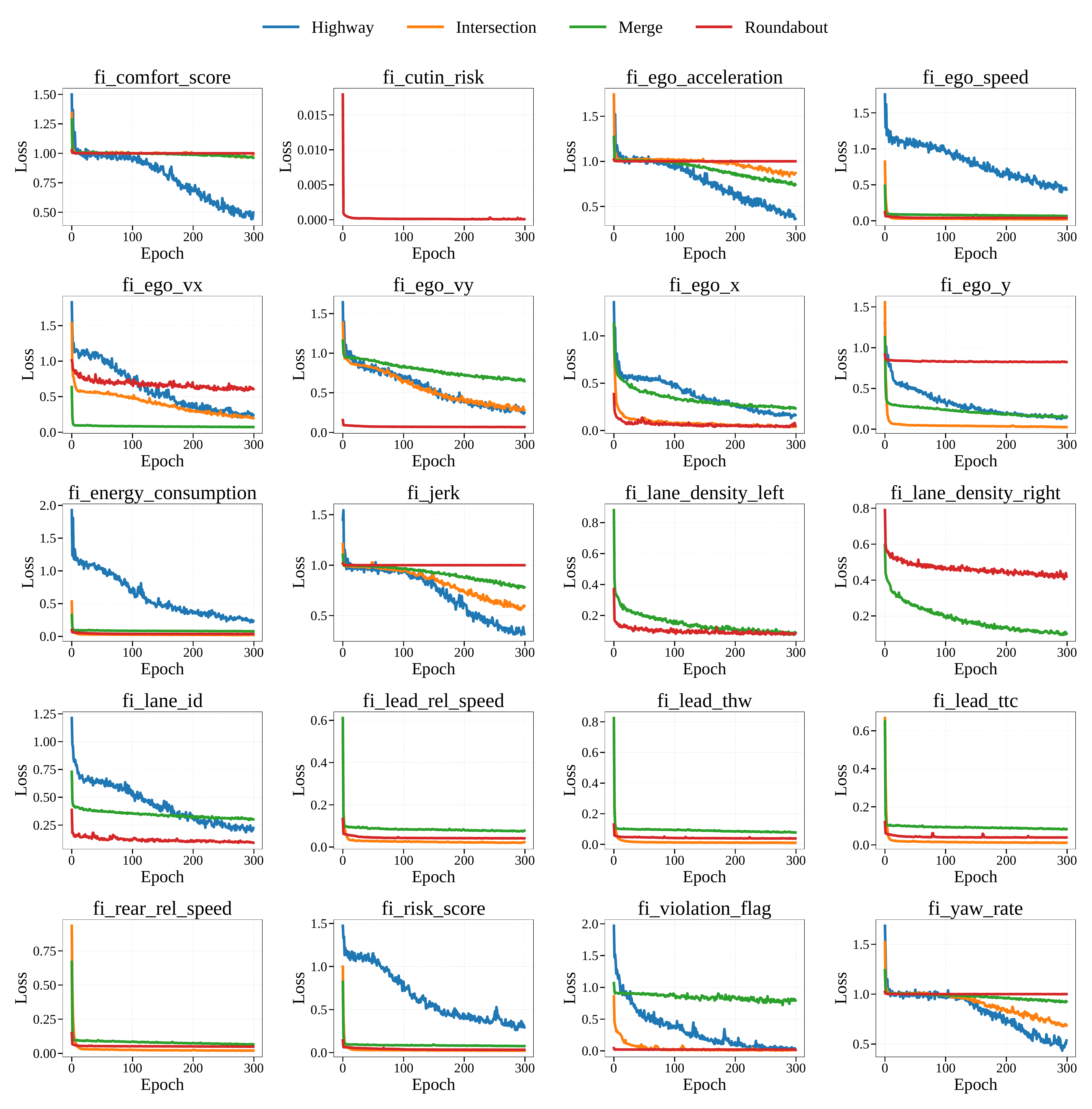}
    \caption{Training loss convergence for the World Dynamics Learning. Each subplot shows one variable's dynamics learning across four scenarios (Highway, Intersection, Merge, Roundabout). All variables converge within 150 epochs, with scenario-dependent convergence rates reflecting inherent complexity differences.}
    \label{fig:training_loss}
\end{figure*}

\subsection{Training Convergence}
\label{app:training_loss}

We train the world dynamics model for 300 epochs using the Adam optimizer with a learning rate of $2 \times 10^{-4}$ and batch size 64. Each variable-specific dynamics function $f_i$ is implemented as a two-layer Transformer with four attention heads. Training is conducted on 50k episodes collected from four scenarios (Highway, Intersection, Merge, and Roundabout) using NVIDIA A100 GPUs. The model learns 20 variable-specific dynamics functions, including ego-state variables (\texttt{ego\_speed}, \texttt{ego\_acceleration}, \texttt{ego\_vx}, \texttt{ego\_vy}, \texttt{ego\_x}, \texttt{ego\_y}, \texttt{yaw\_rate}, \texttt{jerk}, \texttt{lane\_id}), traffic-related variables (\texttt{lead\_ttc}, \texttt{lead\_thw}, \texttt{lead\_rel\_speed}, \texttt{rear\_rel\_speed}, \texttt{lane\_density\_left}, \texttt{lane\_density\_right}, \texttt{cutin\_risk}), and outcome variables (\texttt{risk\_score}, \texttt{comfort\_score}, \texttt{violation\_flag}, \texttt{energy\_consumption}). Figure~\ref{fig:training_loss} shows the training loss curves for all dynamics functions across the four scenarios.

The results suggest that the learned world dynamics are scenario-adaptive not only in difficulty but also in structural composition: different scenarios activate and learn different subsets of variable-specific dynamics functions. Interaction-heavy variables such as \texttt{cutin\_risk}, \texttt{lane\_density\_left/right}, and \texttt{rear\_rel\_speed} are prominent in Merge and Intersection, where lane changes and merging behaviors are frequent, but are weakly activated or effectively absent in Highway, where traffic flow is comparatively stable. Conversely, lateral-motion variables (e.g., \texttt{ego\_vy}, \texttt{yaw\_rate}) become substantially more relevant in Roundabout due to persistent curvature, while contributing marginally in straight-road scenarios. This indicates that the model does not impose a fixed global mechanism set across environments, but instead learns scenario-specific mechanism structures, supporting structural adaptability to heterogeneous traffic contexts.

Beyond these structural differences, the loss curves exhibit consistent convergence patterns. Most variables converge within 100–150 epochs. Simple kinematic variables (e.g., \texttt{ego\_x}, \texttt{ego\_y}, \texttt{lead\_thw}) converge within 50 epochs, whereas interaction-intensive outcome variables (e.g., \texttt{risk\_score}, \texttt{violation\_flag}) require longer training. Highway achieves the lowest final loss due to simpler lane-following dynamics. Roundabout exhibits higher loss for lateral variables because of curved trajectories, while Intersection shows higher variance in \texttt{violation\_flag} due to diverse right-of-way interactions. Safety-related variables (e.g., \texttt{lead\_ttc}, \texttt{cutin\_risk}) converge rapidly in early training when relevant interactions are present, indicating strong dependency signals. In terms of final loss magnitude, continuous variables (e.g., \texttt{ego\_speed}, \texttt{ego\_acceleration}) achieve losses below 0.5, whereas discrete or binary variables (e.g., \texttt{violation\_flag}, \texttt{lane\_id}) stabilize at higher loss due to their classification nature.

\subsection{Trajectory Accuracy}
\label{app:trajectory}

We evaluate trajectory accuracy using Average Displacement Error (ADE), Final Displacement Error (FDE), and cumulative episode reward (Reward) across four driving scenarios: Highway, Merge, Roundabout, and Intersection. ADE and FDE measure the mean and final Euclidean distance between predicted and executed trajectories, respectively, while Reward reflects overall driving efficiency and safety. All methods are evaluated under identical environment settings with the same observation space, action space, and rollout horizon.

Table~\ref{tab:overallresults} reports the mean performance over all evaluation episodes. REINS consistently achieves the best performance across all scenarios in terms of both trajectory accuracy and cumulative reward. In Highway, REINS reduces ADE and FDE to 15.9 and 20.5, respectively, outperforming the strongest baseline (SymAgent) by 12.2\% in ADE and 16.3\% in FDE, while also achieving the highest reward. Similar improvements are observed in Merge and Roundabout, where REINS yields substantial gains in both short-horizon and long-horizon prediction accuracy, indicating more stable multi-step trajectory generation.

The performance gap becomes more pronounced in complex interaction scenarios. In Intersection, REINS achieves an ADE of 5.6 and FDE of 9.7, representing relative reductions of 30.0\% and 13.4\% compared to SymAgent, \rev{and 41.7\% (ADE) and 28.7\% (FDE) compared to MPC}. This suggests that integrating rule-guided action selection with world-model-based forward rollout is particularly effective when decision-making requires anticipating multi-agent interactions and right-of-way constraints.

Across all scenarios, higher trajectory accuracy strongly correlates with higher episode reward, confirming that more accurate future state prediction leads to safer and more efficient planning. Methods relying primarily on language-based reasoning (LLaDA) exhibit the largest errors, especially in high-speed or curved-road settings, highlighting the limitations of planning without explicit dynamics modeling. MPC improves performance by incorporating learned motion predictors, while SymAgent further benefits from symbolic constraints. REINS combines symbolic rule guidance with learned world dynamics, enabling both structured action filtering and predictive verification, which explains its consistent superiority across heterogeneous traffic scenarios.

\begin{table}[!tbp]
\centering
\scriptsize
\setlength{\tabcolsep}{2pt}
{
\fontsize{8.4pt}{10pt}\selectfont
\begin{tabular}{c|c|ccc|c}
\toprule
\textbf{Scen.} & \textbf{Metric}
    & \textbf{LLaDA}
    & \textbf{MPC}
    & \textbf{SymAgent}
    & \textbf{REINS}\\
\midrule
\multirow{3}{*}{Hig.}
& ADE$\downarrow$ & 26.5 & 20.1 & 18.1 & \textbf{15.9} \\
& FDE$\downarrow$ & 29.5 & 25.1 & 24.5 & \textbf{20.5} \\
& Reward$\uparrow$ & 14.7 & 21.6 & 23.1 & \textbf{26.3} \\
\midrule
\multirow{3}{*}{Mer.}
& ADE$\downarrow$ & 17.6 & 13.3 & 12.7 & \textbf{9.9} \\
& FDE$\downarrow$ & 21.8 & 20.0 & 18.5 & \textbf{15.2} \\
& Reward$\uparrow$ & 25.2 & 50.7 & 54.5 & \textbf{58.4} \\
\midrule
\multirow{3}{*}{Rou.}
& ADE$\downarrow$ & 19.7 & 16.5 & 14.3 & \textbf{10.1} \\
& FDE$\downarrow$ & 23.2 & 21.7 & 19.4 & \textbf{14.9} \\
& Reward$\uparrow$ & 12.3 & 22.9 & 23.6 & \textbf{27.8} \\
\midrule
\multirow{3}{*}{Int.}
& ADE$\downarrow$ & 21.6 & 9.6 & 8.0 & \textbf{5.6} \\
& FDE$\downarrow$ & 25.4 & 13.6 & 11.2 & \textbf{9.7} \\
& Reward$\uparrow$ & 35.3 & 54.2 & 56.8 & \textbf{60.5} \\
\bottomrule
\end{tabular}
}
\caption{Trajectory accuracy comparison (mean values).}
\label{tab:overallresults}
\end{table}

\begin{table}[!t]
\centering
\setlength{\tabcolsep}{2pt}
{
\fontsize{8.4pt}{10pt}\selectfont
\begin{tabular}{c|c|c|ccccc}
\toprule
\textbf{Dataset} & \textbf{Scen.} & \textbf{Metric} & $L$=2 & $L$=4 & $L$=\textbf{6} & $L$=8 & $L$=10 \\
\midrule
\multirow{8}{*}{\rotatebox{90}{\textit{Sim.}}}
& \multirow{4}{*}{Hig.}
& Compl. & 0.9934 & 0.9948 & \textbf{0.9956} & 0.9952 & 0.9947 \\
&& RE & -0.42 & 0.31 & \textbf{0.52} & 0.38 & 0.15 \\
&& CR (\%) & 9.34 & 6.89 & \textbf{5.67} & 6.72 & 7.98 \\
&& PE (ms) & \textbf{2.21} & 2.35 & 2.50 & 2.68 & 2.89 \\
\cmidrule{2-8}
& \multirow{4}{*}{Rou.}
& Compl. & 0.9898 & 0.9915 & \textbf{0.9925} & 0.9918 & 0.9908 \\
&& RE & -1.23 & -0.78 & \textbf{-0.54} & -0.72 & -0.95 \\
&& CR (\%) & 14.12 & 10.78 & \textbf{9.01} & 10.34 & 11.89 \\
&& PE (ms) & \textbf{3.18} & 3.41 & 3.62 & 3.89 & 4.21 \\
\midrule
\multirow{8}{*}{\rotatebox{90}{\textit{Real}}}
& \multirow{4}{*}{highD}
& Compl. & 0.9845 & 0.9858 & \textbf{0.9867} & 0.9861 & 0.9852 \\
&& RE & -0.35 & 0.18 & \textbf{0.38} & 0.25 & 0.08 \\
&& CR (\%) & 10.12 & 7.34 & \textbf{6.12} & 7.12 & 8.56 \\
&& PE (ms) & \textbf{2.38} & 2.52 & 2.62 & 2.87 & 3.12 \\
\cmidrule{2-8}
& \multirow{4}{*}{rounD}
& Compl. & 0.9823 & 0.9842 & \textbf{0.9856} & 0.9848 & 0.9834 \\
&& RE & -1.12 & -0.65 & \textbf{-0.41} & -0.58 & -0.82 \\
&& CR (\%) & 15.23 & 11.34 & \textbf{9.45} & 10.78 & 12.45 \\
&& PE (ms) & \textbf{3.52} & 3.76 & 3.89 & 4.25 & 4.63 \\
\bottomrule
\end{tabular}
}
\caption{Sensitivity to temporal lag length $L$ (default: 6).}
\label{tab:sensitivity_lag_app}
\end{table}

\begin{table}[htbp]
\centering
\small
\setlength{\tabcolsep}{2.5pt}
\fontsize{8pt}{9.5pt}\selectfont
\begin{tabular}{l|cccc|cc|r}
\toprule
\multirow{2}{*}{\textbf{Template}} & \multicolumn{4}{c|}{\textit{Simulation}} & \multicolumn{2}{c|}{\textit{Real-world}} & \multirow{2}{*}{\textbf{Total}} \\
 & Hig. & Mer. & Rou. & Int. & highD & rounD & \\
\midrule
\texttt{IF-THEN} & 42 & 48 & 50 & 55 & 38 & 43 & 276 \\
\texttt{AND-THEN} & 38 & 45 & 47 & 52 & 32 & 37 & 251 \\
\texttt{THRESHOLD} & 32 & 35 & 36 & 40 & 30 & 32 & 205 \\
\texttt{PRIORITY} & 12 & 20 & 22 & 28 & 15 & 21 & 118 \\
\texttt{MULTI-ACTION} & 10 & 13 & 14 & 17 & 11 & 12 & 77 \\
\texttt{WEIGHTED-ACTION} & 6 & 8 & 9 & 11 & 7 & 10 & 51 \\
\texttt{TEMPORAL-CHAIN} & 5 & 6 & 7 & 7 & 5 & 15 & 45 \\
\midrule
\textbf{Total} & 145 & 175 & 185 & 210 & 138 & 170 & \textbf{1,023} \\
\bottomrule
\end{tabular}
\caption{\rev{Post-deployment distribution of rules by template type across driving scenarios (after 1000-episode self-evolution per scenario).}}
\label{tab:rule_dist}
\end{table}

\begin{figure}[!tbp]
\centering
\includegraphics[width=0.9\linewidth]{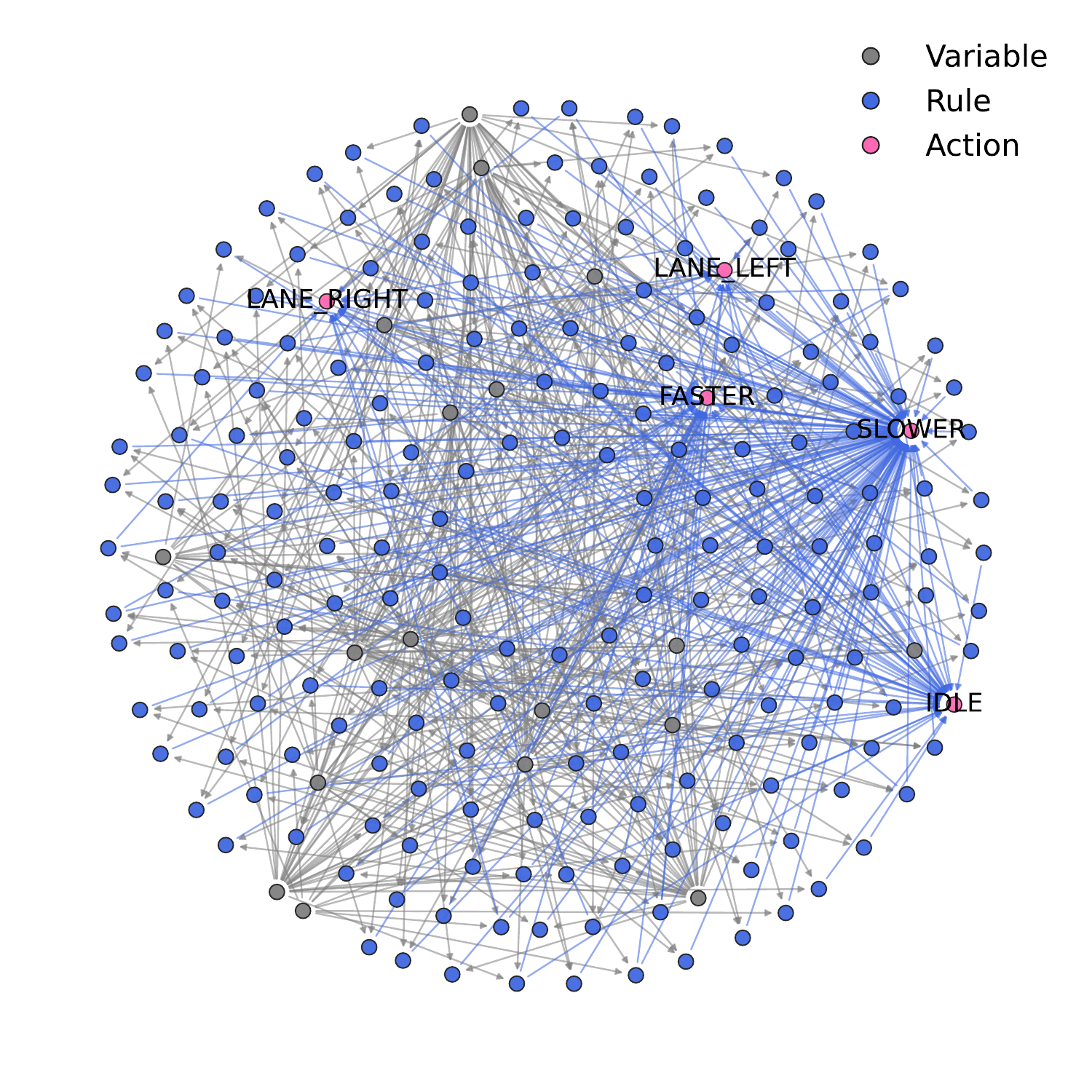}
\caption{Graph structure of 200 randomly sampled rules.}
\label{fig:rule_graph}
\end{figure}

\subsection{Sensitivity to Retrieval Count $k$ and Planning Horizon $H$}
\label{app:sensitivity_kh}

We analyze sensitivity to two runtime hyperparameters: rule retrieval count $k$ (the maximum number of rules returned per scene by the Cypher query) and planning horizon $H$ (the rollout depth used by trajectory verification). Both are evaluated on simulated (Highway, Roundabout) and real-world (highD, rounD) scenarios.

\textbf{Rule Retrieval Count $k$.}
Table~\ref{tab:sensitivity_k} reports results for varying $k$. Small values ($k\!\le\!3$) miss critical rules, increasing CR by 3.5--4.7 points; large values ($k\!\ge\!15$) introduce redundancy, \rev{raising PE by 60--71\% at $k{=}20$} without safety gains. The optimal $k{=}10$ generalizes across domains, with real-world and roundabout scenarios showing higher sensitivity than structured highway settings.

\textbf{Planning Horizon $H$.}
Table~\ref{tab:sensitivity_horizon} reports results for varying $H$. Short horizons ($H\!\le\!3$) cause myopic decisions with 2.9--5.1 point CR increase; \rev{long horizons raise PE by 35--44\% at $H{=}7$ and roughly double it at $H{=}10$}, with marginal safety benefits. The optimal $H{=}5$ balances safety and efficiency, though complex scenarios (Roundabout, rounD) require longer horizons to anticipate multi-agent interactions.

\begin{table}[t]
\centering
\begin{subtable}[t]{0.52\linewidth}
\centering
\resizebox{\linewidth}{!}{%
\setlength{\tabcolsep}{1pt}
\fontsize{8.4pt}{10pt}\selectfont
\begin{tabular}{c|c|c|ccccc}
\toprule
\textbf{Dataset} & \textbf{Scen.} & \textbf{Metric} & $k$=3 & $k$=5 & $k$=\textbf{10} & $k$=15 & $k$=20 \\
\midrule
\multirow{8}{*}{\rotatebox{90}{\textit{Sim.}}}
& \multirow{4}{*}{Hig.}
& Compl. & 0.9712 & 0.9856 & \textbf{0.9956} & 0.9948 & 0.9951 \\
&& RE & 0.18 & 0.41 & \textbf{0.52} & 0.35 & 0.21 \\
&& CR (\%) & 9.21 & 7.08 & \textbf{5.67} & 6.02 & 6.48 \\
&& PE (ms) & \textbf{1.82} & 2.13 & 2.50 & 3.21 & 4.15 \\
\cmidrule{2-8}
& \multirow{4}{*}{Rou.}
& Compl. & 0.9589 & 0.9778 & \textbf{0.9925} & 0.9912 & 0.9908 \\
&& RE & -1.12 & -0.78 & \textbf{-0.54} & -0.72 & -0.89 \\
&& CR (\%) & 13.34 & 10.45 & \textbf{9.01} & 9.62 & 10.28 \\
&& PE (ms) & \textbf{2.67} & 3.12 & 3.62 & 4.56 & 5.78 \\
\midrule
\multirow{8}{*}{\rotatebox{90}{\textit{Real}}}
& \multirow{4}{*}{highD}
& Compl. & 0.9689 & 0.9798 & \textbf{0.9867} & 0.9856 & 0.9852 \\
&& RE & 0.05 & 0.25 & \textbf{0.38} & 0.22 & 0.09 \\
&& CR (\%) & 10.01 & 7.63 & \textbf{6.12} & 6.58 & 7.08 \\
&& PE (ms) & \textbf{2.05} & 2.38 & 2.62 & 3.52 & 4.48 \\
\cmidrule{2-8}
& \multirow{4}{*}{rounD}
& Compl. & 0.9534 & 0.9712 & \textbf{0.9856} & 0.9842 & 0.9836 \\
&& RE & -1.02 & -0.65 & \textbf{-0.41} & -0.58 & -0.76 \\
&& CR (\%) & 14.12 & 11.08 & \textbf{9.45} & 10.02 & 10.75 \\
&& PE (ms) & \textbf{2.98} & 3.48 & 3.89 & 5.02 & 6.28 \\
\bottomrule
\end{tabular}%
}
\caption{Rule retrieval count $k$.}
\label{tab:sensitivity_k}
\end{subtable}
\hfill
\begin{subtable}[t]{0.46\linewidth}
\centering
\resizebox{\linewidth}{!}{%
\setlength{\tabcolsep}{1pt}
\fontsize{8.4pt}{10pt}\selectfont
\begin{tabular}{c|c|c|cccc}
\toprule
\textbf{Dataset} & \textbf{Scen.} & \textbf{Metric} & $H$=3 & $H$=\textbf{5} & $H$=7 & $H$=10 \\
\midrule
\multirow{8}{*}{\rotatebox{90}{\textit{Sim.}}}
& \multirow{4}{*}{Hig.}
& Compl. & 0.9948 & \textbf{0.9956} & 0.9953 & 0.9949 \\
&& RE & -0.12 & \textbf{0.52} & 0.48 & 0.41 \\
&& CR (\%) & 8.56 & \textbf{5.67} & 5.91 & 5.78 \\
&& PE (ms) & \textbf{1.87} & 2.50 & 3.42 & 5.18 \\
\cmidrule{2-7}
& \multirow{4}{*}{Rou.}
& Compl. & 0.9908 & \textbf{0.9925} & 0.9919 & 0.9912 \\
&& RE & -1.02 & \textbf{-0.54} & -0.62 & -0.71 \\
&& CR (\%) & 12.78 & \textbf{9.01} & 9.62 & 9.38 \\
&& PE (ms) & \textbf{2.78} & 3.62 & 4.89 & 7.34 \\
\midrule
\multirow{8}{*}{\rotatebox{90}{\textit{Real}}}
& \multirow{4}{*}{highD}
& Compl. & 0.9856 & \textbf{0.9867} & 0.9862 & 0.9854 \\
&& RE & -0.08 & \textbf{0.38} & 0.34 & 0.28 \\
&& CR (\%) & 9.78 & \textbf{6.12} & 6.48 & 6.32 \\
&& PE (ms) & \textbf{2.12} & 2.62 & 3.76 & 5.62 \\
\cmidrule{2-7}
& \multirow{4}{*}{rounD}
& Compl. & 0.9838 & \textbf{0.9856} & 0.9848 & 0.9841 \\
&& RE & -0.89 & \textbf{-0.41} & -0.52 & -0.62 \\
&& CR (\%) & 14.56 & \textbf{9.45} & 10.06 & 9.78 \\
&& PE (ms) & \textbf{3.15} & 3.89 & 5.52 & 8.18 \\
\bottomrule
\end{tabular}%
}
\caption{Planning horizon $H$.}
\label{tab:sensitivity_horizon}
\end{subtable}
\caption{Sensitivity to retrieval count $k$ (left) and planning horizon $H$ (right).}
\label{tab:sensitivity}
\end{table}

\subsection{Sensitivity to Temporal Lag Length}
\label{app:sensitivity_lag}

The temporal lag length $L$ determines how many historical timesteps the world dynamics model uses when predicting future states. Table~\ref{tab:sensitivity_lag_app} reports results on both simulated and real-world scenarios.

When using small lag lengths ($L \leq 2$), the model fails to capture cumulative temporal patterns such as sustained tailgating or gradual lane drift, leading to substantially higher collision rates. On the simulated Highway scenario, the collision rate increases from 5.67\% at $L=6$ to 9.34\%, corresponding to a 65\% relative increase. A similar degradation is observed on the real-world highD dataset, where the collision rate rises from 6.12\% to 10.12\%. These results indicate that short-term observations are insufficient to infer latent interaction states, such as risk accumulation and driver intent, which are critical for safe planning. In contrast, large lag lengths ($L \geq 10$) introduce excessive historical noise and dilute recent dependency signals, reducing the effective signal-to-noise ratio for short-horizon prediction. Consequently, collision rates increase by approximately 40\% on simulated scenarios compared to the optimal setting. This suggests that overly long histories may impair the model's ability to focus on immediately relevant dynamics, particularly in rapidly evolving traffic situations.

Overall, $L=6$ achieves the best and most consistent performance across both simulated and real-world domains, corresponding to approximately 1.2 seconds of historical context at a sampling rate of 5\,Hz. This time window appears sufficient to capture short-term interaction dynamics while avoiding the accumulation of outdated or irrelevant information, thereby supporting stable multi-step rollout and reliable safety verification in downstream planning.

\subsection{Self-Evolution Across Scenarios}
\label{app:self_evolution_extended}

\begingroup\revcolor
This appendix reports self-evolution dynamics on the remaining simulated scenarios (Merge, Roundabout, Intersection) and the real-world datasets (highD, rounD), complementing the Highway results in Section~\ref{sec:self_evolution}. For each scenario, we deploy REINS for 1000 episodes from the offline-compiled memory $\mathcal{R}_0$ and report $|\mathcal{R}|$, LLM call rate, Compl., and CR at four deployment checkpoints (ep.\ 250, 500, 750, 1000), with a Frozen $\mathcal{R}_0$ row showing the same 1000-episode window run with acquisition disabled; this row exposes $|\mathcal{R}_0|$ and isolates the contribution of online rule writing from any aggregation effect of running more episodes. The per-scenario $|\mathcal{R}|$ at ep.\ 1000 matches the post-deployment totals in Table~\ref{tab:rule_dist}.

Three patterns emerge across the five extended scenarios, mirroring the Highway dynamics. First, $|\mathcal{R}|$ exhibits saturating growth: approximately 44\% of the rule additions (counted from $|\mathcal{R}_0|$ in the Frozen row to $|\mathcal{R}|$ at ep.\ 1000) occur in the first 250 episodes, with diminishing returns thereafter, so the LLM is invoked predominantly for previously unseen scene clusters. Second, the LLM call rate decays monotonically in all scenarios, {from 68--79\% in the first 250-episode window to 12--19\% by ep.\ 1000 (a 4.2--5.5$\times$ reduction)}; scene complexity raises the absolute size of $\mathcal{R}$ {and the residual call rate (highest on Intersection)} but not the shape of the decay. Third, the Frozen $\mathcal{R}_0$ control consistently underperforms the evolving memory at ep.\ 1000 (mean Compl.\ gap $\approx$0.017, mean CR gap $\approx$4.2 points), confirming that the gains attributable to online acquisition are not artifacts of statistical aggregation over a longer evaluation window. Intersection benefits most (Compl.\ $+0.0218$, CR $-4.87$ points from Frozen to ep.\ 1000), reflecting larger room for online acquisition where the offline $\mathcal{R}_0$ is most incomplete relative to right-of-way reasoning; the other four scenarios cluster in a tight range (Compl.\ $+0.0154$ to $+0.0164$, CR $-3.79$ to $-4.40$ points). highD mirrors Highway and rounD mirrors Roundabout in both $|\mathcal{R}|$ growth shape and Compl./CR convergence, indicating that the harness operates on factored variable signatures rather than scenario-specific surface features. Cross-domain transfer of the compiled memory itself is studied separately in Appendix~\ref{app:transfer}.

\endgroup
\begin{table*}[h]
\centering
\setlength{\tabcolsep}{3pt}
\fontsize{8.4pt}{10pt}\selectfont
\begin{tabular}{l|l|c|c|cc}
\toprule
\textbf{Scenario} & \textbf{Checkpoint} & $|\mathcal{R}|$ & \textbf{LLM Call (\%)} & \textbf{Compl.} ($\uparrow$) & \textbf{CR (\%)} ($\downarrow$) \\
\midrule
\multirow{5}{*}{Merge}
  & ep.\ 250 & 136 & {71.36} & 0.9858 & 8.20 \\
  & ep.\ 500 & 153 & {49.12} & 0.9893 & 7.35 \\
  & ep.\ 750 & 167 & {26.48} & 0.9912 & 7.05 \\
  & ep.\ 1000 (final) & \textbf{175} & \textbf{{13.92}} & \textbf{0.9924} & \textbf{6.86} \\
  & \textit{Frozen $\mathcal{R}_0$} (1000-ep eval) & 105 & --- & 0.9770 & 10.65 \\
\midrule
\multirow{5}{*}{Roundabout}
  & ep.\ 250 & 144 & {73.84} & 0.9852 & 10.55 \\
  & ep.\ 500 & 162 & {52.20} & 0.9888 & 9.85 \\
  & ep.\ 750 & 176 & {28.36} & 0.9910 & 9.40 \\
  & ep.\ 1000 (final) & \textbf{185} & \textbf{{15.24}} & \textbf{0.9925} & \textbf{9.01} \\
  & \textit{Frozen $\mathcal{R}_0$} (1000-ep eval) & 112 & --- & 0.9770 & 13.40 \\
\midrule
\multirow{5}{*}{Intersection}
  & ep.\ 250 & 164 & {78.52} & 0.9742 & 10.85 \\
  & ep.\ 500 & 184 & {57.64} & 0.9789 & 9.95 \\
  & ep.\ 750 & 200 & {33.08} & 0.9817 & 9.45 \\
  & ep.\ 1000 (final) & \textbf{210} & \textbf{{18.76}} & \textbf{0.9838} & \textbf{9.08} \\
  & \textit{Frozen $\mathcal{R}_0$} (1000-ep eval) & 127 & --- & 0.9620 & 13.95 \\
\midrule
\multirow{5}{*}{highD}
  & ep.\ 250 & 107 & {68.24} & 0.9803 & 7.45 \\
  & ep.\ 500 & 121 & {47.56} & 0.9838 & 6.65 \\
  & ep.\ 750 & 131 & {24.12} & 0.9856 & 6.32 \\
  & ep.\ 1000 (final) & \textbf{138} & \textbf{{12.36}} & \textbf{0.9867} & \textbf{6.12} \\
  & \textit{Frozen $\mathcal{R}_0$} (1000-ep eval) & 84 & --- & 0.9712 & 9.95 \\
\midrule
\multirow{5}{*}{rounD}
  & ep.\ 250 & 132 & {74.68} & 0.9785 & 10.95 \\
  & ep.\ 500 & 149 & {53.40} & 0.9821 & 10.20 \\
  & ep.\ 750 & 162 & {29.72} & 0.9842 & 9.78 \\
  & ep.\ 1000 (final) & \textbf{170} & \textbf{{16.08}} & \textbf{0.9856} & \textbf{9.45} \\
  & \textit{Frozen $\mathcal{R}_0$} (1000-ep eval) & 103 & --- & 0.9692 & 13.85 \\
\bottomrule
\end{tabular}
\caption{Self-evolution dynamics across the remaining evaluation scenarios. Same metrics and checkpoints as Table~\ref{tab:self_evolution}; the Frozen row runs the same 1000-episode window with acquisition disabled and exposes $|\mathcal{R}_0|$.}
\label{tab:self_evolution_extended}
\end{table*}

\subsection{Case Study}
\label{app:case_study}

\begin{figure}[!tbp]
\centering
\includegraphics[width=\linewidth]{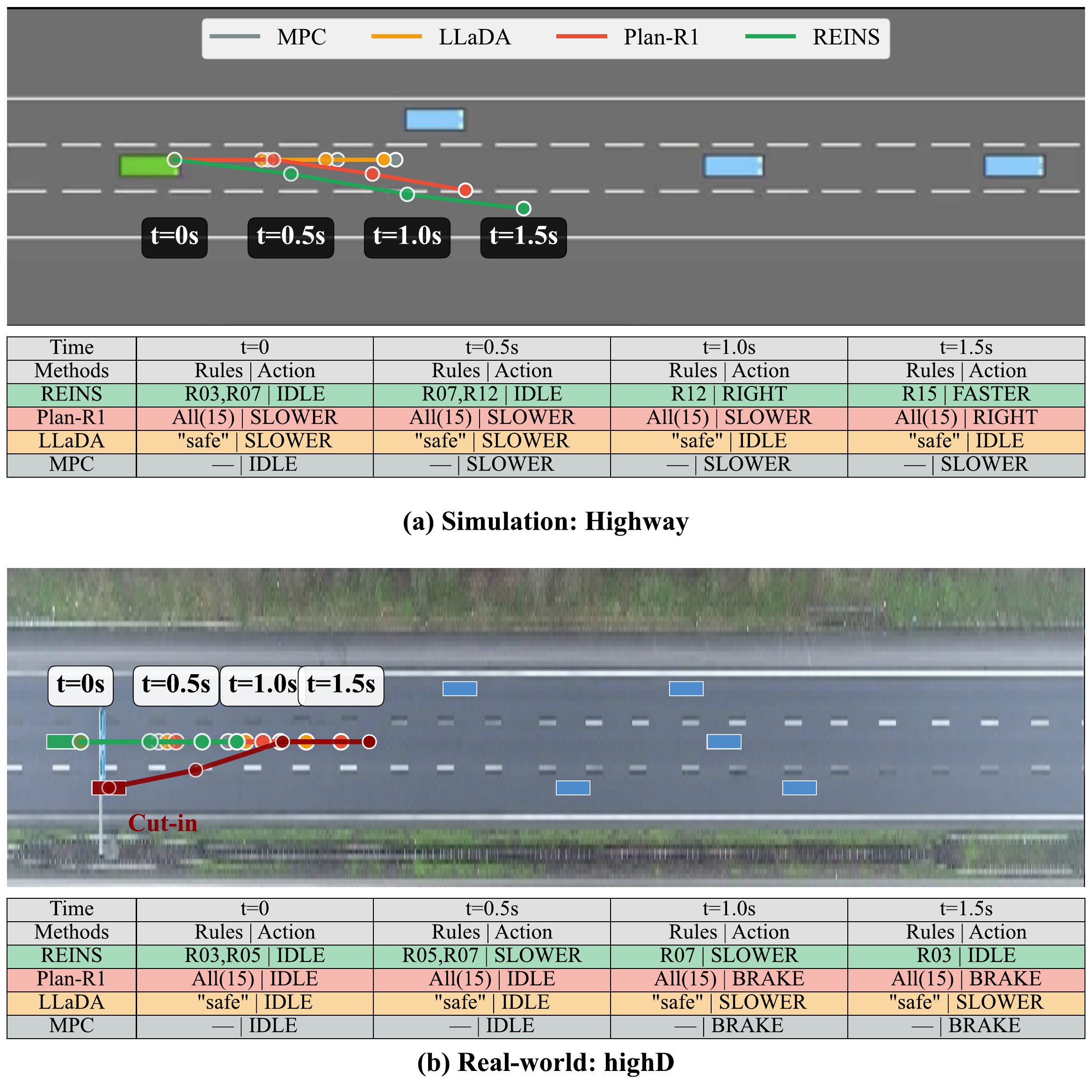}
\caption{Trajectory comparison across four methods on (a) simulated highway and (b) real-world highD scenarios.}
\label{fig:case_study}
\end{figure}

Figure~\ref{fig:case_study} compares trajectories of REINS, Plan-R1, LLaDA, and MPC on two scenarios, with activated rules and actions shown at each timestep. Rule definitions are provided in Appendix~\ref{app:representative_rules}.

\textbf{Simulated Highway.} REINS activates R03 and R07 at $t{=}0$ to monitor the lead vehicle and adjacent lanes, triggers R12 at $t{=}0.5$s when a safe gap appears, executes a lane change at $t{=}1.0$s, and accelerates via R15 at $t{=}1.5$s, producing a smooth action sequence (IDLE$\!\to\!$IDLE$\!\to\!$RIGHT$\!\to\!$FASTER). Plan-R1 applies all 15 rules simultaneously, causing conflicting constraints and repeated deceleration (SLOWER$\!\times\!3$) before a delayed maneuver.

\textbf{Real-World highD.} In the cut-in scenario, REINS detects lateral intrusion at $t{=}0$ via R05, performs preemptive deceleration at $t{=}0.5$s using R07, and maintains safe distance until $t{=}1.5$s (IDLE$\!\to\!$SLOWER$\!\to\!$SLOWER$\!\to\!$IDLE). Plan-R1 responds later with emergency braking at $t{=}1.0$s due to uniform rule execution.

\subsection{Compiled Rule Database}
\label{app:case_study_rules}

\begin{table*}[!htbp]
\centering
\setlength{\tabcolsep}{2pt}
{
% \fontsize{8.4pt}{10pt}\selectfont
\small
\begin{tabular}{c|l|l|c}
\toprule
\textbf{ID} & \textbf{Guard Condition $\mathcal{P}(r)$} & \textbf{Action $a_r$} & \textbf{Template} \\
\midrule
R01 & $\text{TTC} < 2.0\text{s}$ & emergency brake & \texttt{THRESHOLD} \\
R02 & $v_{ego} > v_{limit} + 5\text{km/h}$ & DEC to limit & \texttt{THRESHOLD} \\
R03 & $\text{THW} < 1.8\text{s} \lor \text{DHW} < 25\text{m}$ & DEC / maintain gap & \texttt{IF-THEN} \\
R04 & $\text{TTC} < 3.5\text{s} \land \text{DHW} < 40\text{m}$ & DEC (moderate) & \texttt{AND-THEN} \\
R05 & $v^{lat}_{adj} > 0.3\text{m/s} \land \Delta y < 2\text{m}$ & DEC (preemptive) & \texttt{AND-THEN} \\
R06 & $\text{visibility} < 100\text{m}$ & reduce speed 20\% & \texttt{THRESHOLD} \\
R07 & $\text{TTC}_{adj} < 4\text{s} \lor \text{cut-in}$ & increase headway & \texttt{IF-THEN} \\
R08 & $\text{lane\_end} < 200\text{m}$ & prepare merge & \texttt{TEMPORAL-CHAIN} \\
R09 & $\text{at\_yield} \land \text{traffic\_present}$ & yield / stop & \texttt{PRIORITY} \\
R10 & $\text{pedestrian} < 30\text{m}$ & DEC / prepare stop & \texttt{MULTI-ACTION} \\
R11 & $v_{lead} < v_{ego} - 10\text{km/h} \land \text{gap}_{adj} > 35\text{m}$ & consider lane change & \texttt{AND-THEN} \\
R12 & $\text{gap}_{target} > 30\text{m} \land \text{TTC}_{target} > 5\text{s}$ & execute lane change & \texttt{AND-THEN} \\
R13 & $\text{in\_lane\_change} \land \text{TTC}_{new} < 3\text{s}$ & abort lane change & \texttt{AND-THEN} \\
R14 & $\text{merge\_complete} \land \text{DHW} < 20\text{m}$ & DEC post-merge & \texttt{AND-THEN} \\
R15 & $\text{post\_merge} \land v_{ego} < v_{flow} - 5\text{km/h}$ & ACC to flow & \texttt{AND-THEN} \\
\bottomrule
\end{tabular}
}
\caption{Representative rules from the compiled rule database.}
\label{tab:all_rules}
\end{table*}

\subsubsection{Rule Database Statistics}
\label{app:rule_database_statistics}
Table~\ref{tab:rule_dist} shows the distribution of variables by macro-group and rules by template type in REINS's Neo4j graph database. The post-deployment database contains 1,023 rules covering six driving scenarios (four simulated: Highway, Merge, Roundabout, Intersection; two real-world: highD, rounD). IF-THEN and AND-THEN templates dominate due to the prevalence of conditional and multi-condition reasoning in driving. Intersection scenarios contain the most rules due to complex right-of-way interactions, while Highway scenarios require fewer given simpler lane-following dynamics. This scale is comparable to industrial rule-based systems while remaining tractable for $O(\log n)$ graph database retrieval ($<$1\,ms per query).

Figure~\ref{fig:rule_graph} visualizes the graph structure of 200 randomly sampled rules from the database. Gray nodes are variables, blue nodes are rules, and pink nodes are the five discrete actions. Edge colors match source node colors: gray edges represent Variable$\to$Rule, while blue edges represent Rule$\to$Action.

Several structural patterns emerge from this visualization:
(1) \textbf{Variable sharing}: A small set of variables (e.g., TTC, THW, speed) triggers a large number of rules, indicating that safety-critical variables serve as common preconditions across diverse driving scenarios.
(2) \textbf{Action concentration}: The majority of rules converge to \texttt{SLOWER} and \texttt{IDLE}, which exhibit the highest in-degree among all five actions. This reflects the safety-first design principle where conservative actions are recommended under uncertain or risky conditions. In contrast, \texttt{LANE\_LEFT}, \texttt{LANE\_RIGHT}, and \texttt{FASTER} receive fewer rule connections, as these actions require more stringent preconditions.
(3) \textbf{Rule diversity}: Despite action concentration, rules maintain diverse trigger conditions from different variables, enabling fine-grained context-aware activation rather than uniform rule application.
These patterns support REINS's retrieval strategy: variable-indexed queries efficiently activate relevant rule subsets, while action aggregation via noisy-OR naturally prioritizes safety-critical recommendations.

\subsubsection{Representative Rules}
\label{app:representative_rules}
Table~\ref{tab:all_rules} lists 15 representative rules from REINS's compiled rule database. The full post-deployment Neo4j database contains 1,023 rules across six driving scenarios, from which context-relevant subsets are retrieved at runtime via parameterized Cypher queries. These rules are automatically compiled from LLM-generated linguistic descriptions and calibrated using the threshold calibration mechanism described in Section~\ref{sec:rule_compilation}. For fair comparison, Plan-R1 applies the same 15 representative rules uniformly at each step, while REINS selectively activates 2--3 context-relevant rules via the retrieval mechanism.

The 15 rules cover six of the seven template types defined in Definition~\ref{def:rule_template}:
\textbf{(1) \texttt{THRESHOLD}} rules (R01, R02, R06) enforce numeric boundaries on physical variables such as TTC and speed;
\textbf{(2) \texttt{IF-THEN}} rules (R03, R07) implement basic conditional logic with disjunctive guards;
\textbf{(3) \texttt{AND-THEN}} rules (R04, R05, R11--R15) require multiple conditions to be satisfied simultaneously;
\textbf{(4) \texttt{PRIORITY}} rules (R09) establish decision hierarchies for right-of-way scenarios;
\textbf{(5) \texttt{MULTI-ACTION}} rules (R10) trigger compound actions under critical conditions;
\textbf{(6) \texttt{TEMPORAL-CHAIN}} rules (R08) encode time-sensitive constraints for upcoming events.
The complete definitions of all seven rule template types are provided in Appendix~\ref{app:rules}, and the conflict resolution mechanism is described in Appendix~\ref{app:conflict}.

%%%%%%%%%%%%%%%%%%%%%%%%%%%%%%%%%%%%%%%%%%%%%%%%%%%%%%%%%%%%%%%%%%%%%%%%%%%%%%%
% Appendix D: Fairness Protocol, Robustness, Transfer, and Scalability
% (studies added in response to the NeurIPS 2026 reviews)
%%%%%%%%%%%%%%%%%%%%%%%%%%%%%%%%%%%%%%%%%%%%%%%%%%%%%%%%%%%%%%%%%%%%%%%%%%%%%%%
\begingroup\revcolor
\section{Fairness Protocol, Robustness, Transfer, and Scalability}
\label{app:review}

This appendix collects the studies that scope the harness claim and stress the main results: the domain-specific versus domain-agnostic components of the harness (\ref{app:scope}), the baseline fairness protocol with a matched re-run and a common-reference compliance audit (\ref{app:fairness}), the runtime profile of the PDDL baseline (\ref{app:pddl_profile}), robustness to the LLM backbone and to prompt variation (\ref{app:robustness_llm}), what rollout verification filters (\ref{app:verification_residual}), reactive replay and observation noise (\ref{app:reactive_noise}), cross-domain memory transfer (\ref{app:transfer}), the latency of the online-miss cycle and its scheduling (\ref{app:miss_cycle}), and retrieval at deployment scale (\ref{app:scalability}). Unless stated otherwise, all studies use the Highway scenario, five seeds, and the protocol of Section~\ref{sec:exp_setting}.

\subsection{Scope of the Harness Claim}
\label{app:scope}

The three couplings are presented as a harness pattern, while the implementation is specialized to driving. Table~\ref{tab:scope} classifies each component of REINS and sketches the instantiation for a different trajectory-planning domain, robot-arm manipulation, whose action space is continuous and high-dimensional and whose safety notions (contact force, grasp stability) differ from driving. Two components are domain schemas that any instantiation must author; the remaining five transfer unchanged. The safety notion enters only through the variable schema and the terms of the objective $J$: calibrated predicates over contact-force variables are structurally identical to TTC guards, and the calibration procedure and the verification protocol do not change. The harness contribution lives in the five unchanged components; the two redesigned ones constitute the domain description that any planner for that domain must author anyway.

\begin{table}[htbp]
\centering
\fontsize{8.4pt}{10pt}\selectfont
\setlength{\tabcolsep}{4pt}
\begin{tabular}{>{\raggedright\arraybackslash}p{2.6cm}|>{\raggedright\arraybackslash}p{3.6cm}|>{\raggedright\arraybackslash}p{3.9cm}|>{\raggedright\arraybackslash}p{2.5cm}}
\toprule
\textbf{Component} & \textbf{Driving (this paper)} & \textbf{Manipulation (sketch)} & \textbf{Transfer} \\
\midrule
Variable space and groups & 44 variables in Env, Traffic, Ego, Action, Outcome & joint angles and torques, end-effector pose, contact forces, object poses & redesign \\
Action space & 5 driving meta-actions & skill primitives (reach, grasp, lift, place, retreat) & redesign \\
Predicate grammar and thresholds & TTC and THW guards & force-margin, slip, and grasp-stability guards & grammar unchanged, thresholds recalibrated \\
Dynamics learning & LLM prior, group-weighted ridge, per-variable Transformer & same, retrained on domain data & unchanged \\
Verification & collision and violation terms in $J$ & contact-force and stability terms in $J$ & protocol unchanged \\
Indexed memory and retrieval & variable-keyed graph schema & same & unchanged \\
Acquisition loop & miss detection, proposal, verification, write-back & same & unchanged \\
\bottomrule
\end{tabular}
\caption{Domain-specific and domain-agnostic components of the harness.}
\label{tab:scope}
\end{table}

\subsection{Baseline Fairness Protocol}
\label{app:fairness}

\textbf{Shared and unshared conditions.} PDDL, DriReg, and LLaDA were conceived for symbolic planning domains, regulatory text, and driver handbooks respectively, none with a factored driving state, so each was re-implemented faithfully to its original design for closed-loop highway-env. Table~\ref{tab:fairness_conditions} lists what is shared with REINS and what each method keeps by design. The shared conditions fix the observation space, action space, episodes, LLM backbone, and decoding; the unshared ones leave each method its own knowledge representation and thresholds, which leaves the rule-generation procedure as a confound. Three instruments below isolate the harness regardless.

\begin{table}[htbp]
\centering
\fontsize{8.4pt}{10pt}\selectfont
\setlength{\tabcolsep}{4pt}
\begin{tabular}{>{\raggedright\arraybackslash}p{5.2cm}|c|c|c}
\toprule
\textbf{Condition} & \textbf{PDDL} & \textbf{DriReg} & \textbf{LLaDA} \\
\midrule
Observation space, action space, episodes & shared & shared & shared \\
GPT-4o backbone (2024-08-06), temperature 0 for rule generation & shared & shared & shared \\
44-variable vocabulary (Appendix~\ref{app:variables}) in rule-generation prompts & own vocabulary & own vocabulary & n/a (linguistic rules) \\
Threshold calibration & uncalibrated & uncalibrated & n/a (no thresholds) \\
\bottomrule
\end{tabular}
\caption{Fairness conditions for the rule-based baselines as published. ``Own vocabulary'' and ``uncalibrated'' follow each method's original design.}
\label{tab:fairness_conditions}
\end{table}

\textbf{Matched re-run.} We regenerated the PDDL and DriReg rules over the same 44-variable vocabulary, passed their thresholds through REINS's calibration pipeline, and re-evaluated (Table~\ref{tab:matched_rerun}). Input normalization improves both baselines by 3--5 CR points and explains a minority of the gap; the majority survives it.

\begin{table}[htbp]
\centering
\fontsize{8.4pt}{10pt}\selectfont
\setlength{\tabcolsep}{5pt}
\begin{tabular}{l|cc|cc|c}
\toprule
\textbf{CR (\%)} & \textbf{PDDL} & \textbf{PDDL matched} & \textbf{DriReg} & \textbf{DriReg matched} & \textbf{REINS} \\
\midrule
Highway      & 25.93 & 21.37 & 21.76 & 18.92 & \textbf{5.67} \\
Intersection & 32.12 & 27.64 & 29.89 & 25.81 & \textbf{9.08} \\
\bottomrule
\end{tabular}
\caption{Matched re-run: PDDL and DriReg rules regenerated over REINS's variable vocabulary with calibrated thresholds.}
\label{tab:matched_rerun}
\end{table}

\textbf{Bracketing with the ablations.} Two ablations from Table~\ref{tab:ablation} bracket the remaining gap with the paper's own numbers. Hand-crafted rules inside the full harness still reach CR 11.23\% on Highway, better than every matched no-harness variant in Table~\ref{tab:matched_rerun} and close to the best baseline Plan-R1 at 12.83\%. Conversely, applying REINS's rules uniformly, the way the baselines use rules, raises CR by 22--43\% and PE by 34--58\%. Worse inputs with the harness keep most of the gain; the same inputs without the harness lose it.

\textbf{Common-reference compliance.} Compliance in Table~\ref{tab:mainresults} is measured against each method's own rule set. Table~\ref{tab:common_compliance} re-scores every method against REINS's compiled rule set as a common reference. Compliance is a property of the executed trajectory rather than of the planner: at each step, the reference rules whose guards hold in the logged state become active, and a violation is counted when the logged behavior fails an active rule (Appendix~\ref{app:metrics}). Each method therefore plans exactly as before and its trajectories are audited afterward, which is also how the four methods without native rule sets receive values. Because the reference set equals REINS's planning set, REINS's own value is unchanged by construction; the informative content is that the baseline ordering and the gaps persist.

\begin{table}[htbp]
\centering
\fontsize{8.4pt}{10pt}\selectfont
\setlength{\tabcolsep}{5pt}
\begin{tabular}{l|cc|cc}
\toprule
\multirow{2}{*}{\textbf{Method}} & \multicolumn{2}{c|}{\textbf{Own rules}} & \multicolumn{2}{c}{\textbf{Common reference (REINS set)}} \\
 & Highway & Intersection & Highway & Intersection \\
\midrule
MPC        & n/a    & n/a    & 0.8712 & 0.8423 \\
MCTS       & n/a    & n/a    & 0.8831 & 0.8517 \\
DriveSpeak & n/a    & n/a    & 0.8246 & 0.7912 \\
LLaDA      & n/a    & n/a    & 0.8412 & 0.8104 \\
PDDL       & 0.9123 & 0.8789 & 0.8867 & 0.8621 \\
DriReg     & 0.9452 & 0.9153 & 0.9083 & 0.8874 \\
SymAgent   & 0.9561 & 0.9274 & 0.9312 & 0.9067 \\
Plan-R1    & 0.9717 & 0.9553 & 0.9498 & 0.9231 \\
REINS      & \textbf{0.9956} & \textbf{0.9838} & \textbf{0.9956} & \textbf{0.9838} \\
\bottomrule
\end{tabular}
\caption{Compliance against each method's own rule set and against REINS's compiled rule set as a common reference.}
\label{tab:common_compliance}
\end{table}

\subsection{Runtime Profile of the PDDL Baseline}
\label{app:pddl_profile}

The PDDL baseline follows the pipeline in which the LLM writes the domain's preconditions and effects offline and a classical solver executes at runtime~\citep{guan2023leveraging,liu2023llmp}. The solver is Pyperplan~\citep{alkhazraji2020pyperplan}, a Python STRIPS planner invoked in process at each decision step; there is no per-step LLM call. Its 4.95\,ms per step on Highway (Table~\ref{tab:mainresults}) decomposes into grounding 3.71\,ms, search 0.94\,ms, and plan extraction 0.30\,ms. The solver re-instantiates every precondition against the current state at every step with no indexing, so grounding takes roughly three quarters of the budget. REINS avoids exactly this cost: variable-indexed retrieval touches 2--3 rules in under 1\,ms, followed by a fixed-size rollout at $H{=}5$. This is the architectural payoff of consolidation with indexing.

\subsection{Robustness to the LLM Backbone and to Prompt Variation}
\label{app:robustness_llm}

The LLM produces the dependency priors, the rule structures, and the Cypher queries, so we regenerate the whole offline pipeline with two further backbones and with three paraphrases of the dependency and rule-extraction prompts (Table~\ref{tab:backbone}). Rule structure varies mildly across backbones and paraphrases, while the calibrated thresholds and the final behavior barely move. The LLM picks variables, templates, and actions; calibration re-anchors every numeric threshold to data, so backbone differences wash out at the threshold level. Malformed or unsafe proposals never reach memory, because rollout verification rejects them regardless of which backbone or prompt produced them.

\begin{table}[htbp]
\centering
\fontsize{8.4pt}{10pt}\selectfont
\setlength{\tabcolsep}{5pt}
\begin{tabular}{l|c|c|c|c}
\toprule
\textbf{Variation (Highway)} & \textbf{\begin{tabular}[c]{@{}c@{}}Rule overlap\\with GPT-4o\end{tabular}} & \textbf{\begin{tabular}[c]{@{}c@{}}Threshold shift\\after calibration\end{tabular}} & \textbf{Compl.} & \textbf{CR (\%)} \\
\midrule
GPT-4o (paper)            & n/a          & n/a   & 0.9956 & 5.67 \\
Claude Sonnet 4.5         & 87.3\%       & 3.2\% & 0.9948 & 5.89 \\
Qwen2.5-72B-Instruct      & 81.6\%       & 3.8\% & 0.9931 & 6.24 \\
\midrule
Three prompt paraphrases, GPT-4o & 91.2--94.6\% & --- & --- & 5.58--5.91 \\
\bottomrule
\end{tabular}
\caption{Robustness of the compiled rules and of closed-loop behavior to the LLM backbone and to prompt paraphrases.}
\label{tab:backbone}
\end{table}

\subsection{What Rollout Verification Filters}
\label{app:verification_residual}

On the 1000-episode Highway deployment of Section~\ref{sec:self_evolution}, applying the windowed call rates of Table~\ref{tab:self_evolution} to the 100{,}000 deployment steps yields roughly 39{,}000 logged proposals (an arithmetic consequence of the call rates, not an independent count). Of these, 2.1\% failed schema validation, 18.0\% were rejected by rollout verification, and 79.9\% were accepted. Acceptances overwhelmingly hash-collide with rules already in memory (Eq.~\ref{eq:memory_update}) and only update confidences; they deduplicate to the 57 unique rules that grow the memory from 88 to 145. A manual audit of the 57 unique rules judged 54 consistent with domain guidelines; the three flagged rules were over-specific and never activated again. Episodes in which at least one acquired rule fired show no elevated collision rate against episodes in which none fired (5.4\% against 5.8\%).

Three properties of the design bound what a plausible but incorrect proposal can do. Verification errs toward rejection: a rule is accepted only if every calibrated predicate holds along the predicted horizon (Eq.~\ref{eq:online_verify}), so dynamics error costs availability rather than safety in the common case. Accepted rules never bypass the runtime loop: every executed action is re-verified by rollout at decision time (Section~\ref{sec:rule_inference}), so a wrongly accepted rule must also survive the per-step check before it can affect control. Finally, the residual dependence on dynamics fidelity is stated in Section~\ref{sec:limitations}.

\subsection{Reactive Replay and Observation Noise}
\label{app:reactive_noise}

\textbf{Reactive replay.} In the four simulated scenarios the surrounding vehicles already react to the ego through IDM car-following with MOBIL lane changes. The real-world rows keep highD and rounD as recorded so that they reflect human behavior rather than a simulator's. To measure the effect of this choice, a reactive-replay variant on highD switches recorded vehicles within 50\,m of the ego to IDM once interaction occurs, following the reactive closed-loop protocol of nuPlan~\citep{caesar2021nuplan}. REINS CR moves from 6.12\% to 6.94\% while Plan-R1 moves from 12.78\% to 15.36\%, so the gap to the strongest baseline widens.

\textbf{Observation noise.} We corrupt the observed variables with zero-mean Gaussian noise scaled to each variable's standard deviation and, separately, drop 5\% of frames (Table~\ref{tab:noise}). Calibrated thresholds carry a data-derived margin and every action is vetted by a multi-step rollout rather than by a single reading, so REINS degrades more slowly than the strongest baseline at every level, with Compl.\ staying at or above 0.987. Full-stack perception errors (occlusion, tracking identity switches, map errors) remain out of scope.

\begin{table}[htbp]
\centering
\fontsize{8.4pt}{10pt}\selectfont
\setlength{\tabcolsep}{6pt}
\begin{tabular}{l|c|ccc|c}
\toprule
\textbf{CR (\%) on Highway} & \textbf{no noise} & \textbf{2.5\% noise} & \textbf{5\% noise} & \textbf{10\% noise} & \textbf{5\% frame drops} \\
\midrule
REINS   & 5.67  & 5.92  & 6.51  & 8.14  & 6.88 \\
Plan-R1 & 12.83 & 13.71 & 15.42 & 19.87 & 16.50 \\
\bottomrule
\end{tabular}
\caption{Collision rate under $\sigma$-scaled Gaussian observation noise and under frame drops.}
\label{tab:noise}
\end{table}

\subsection{Cross-Domain Memory Transfer}
\label{app:transfer}

We test how the memory behaves when the offline data are domain-shifted, using the checkpoint protocol of Table~\ref{tab:self_evolution}. Setting A compiles $\mathcal{R}_0$ solely from highD and rounD (whose memories total 187 rules before merging and content-hash deduplication) and deploys on highway-env; Setting B compiles it solely from the four simulated scenarios (432 rules before deduplication) and deploys on highD and rounD. Table~\ref{tab:transfer} reports the ep.\ 250 and ep.\ 1000 endpoints of each setting, together with a frozen-transfer control on Highway; the native reference rows come from Table~\ref{tab:self_evolution} and Table~\ref{tab:self_evolution_extended}.

\begin{table}[htbp]
\centering
\fontsize{8.4pt}{10pt}\selectfont
\setlength{\tabcolsep}{5pt}
\begin{tabular}{l|l|c|c|c|c|c}
\toprule
\textbf{Scenario} & \textbf{Checkpoint} & $|\mathcal{R}|$ & \textbf{LLM Call (\%)} & \textbf{Compl.} & \textbf{CR (\%)} & \textbf{PE (ms)} \\
\midrule
\multicolumn{7}{l}{\textit{Setting A: $\mathcal{R}_0$ from highD and rounD, deployed on highway-env}} \\
\midrule
\multirow{4}{*}{Highway}
 & ep.\ 250            & 265 & 86.2 & 0.9612 & 11.84 & 2.71 \\
 & ep.\ 1000           & 349 & 14.9 & 0.9914 & 6.21  & 2.54 \\
 & ep.\ 1000, acquisition frozen & --- & --- & --- & 11.20 & --- \\
 & native, ep.\ 1000   & 145 & 12.81 & 0.9956 & 5.67 & 2.50 \\
\midrule
\multirow{3}{*}{Intersection}
 & ep.\ 250            & 287 & 91.3 & 0.9487 & 15.62 & 4.87 \\
 & ep.\ 1000           & 391 & 17.2 & 0.9782 & 10.13 & 4.66 \\
 & native, ep.\ 1000   & 210 & 12.81 & 0.9838 & 9.08 & 4.58 \\
\midrule
\multicolumn{7}{l}{\textit{Setting B: $\mathcal{R}_0$ from the four simulated scenarios, deployed on highD and rounD}} \\
\midrule
\multirow{3}{*}{highD}
 & ep.\ 250            & 471 & 71.4 & 0.9702 & 9.87  & 2.79 \\
 & ep.\ 1000           & 508 & 13.6 & 0.9838 & 6.79  & 2.68 \\
 & native, ep.\ 1000   & 138 & 12.81 & 0.9867 & 6.12 & 2.62 \\
\midrule
\multirow{3}{*}{rounD}
 & ep.\ 250            & 478 & 79.8 & 0.9645 & 14.21 & 4.12 \\
 & ep.\ 1000           & 517 & 15.1 & 0.9821 & 10.24 & 3.97 \\
 & native, ep.\ 1000   & 170 & 12.81 & 0.9856 & 9.45 & 3.89 \\
\bottomrule
\end{tabular}
\caption{Cross-domain memory transfer under the checkpoint protocol of Table~\ref{tab:self_evolution}. The frozen-transfer row disables acquisition after transfer; only its CR was recorded.}
\label{tab:transfer}
\end{table}

A domain-shifted memory starts degraded: at ep.\ 250 the call rate runs 86.2\% on Highway against the native 70.0\% of Table~\ref{tab:self_evolution}, and CR runs several points above the native value at the same checkpoint (11.84\% against 7.05\%). Acquisition recovers most of the gap within 1000 episodes: final CR lands within 0.5--1.1 points of native and Compl.\ within 0.003--0.006. On Highway, freezing the transferred memory prevents recovery: at ep.\ 1000 the frozen transfer stays at CR 11.20\% against 6.21\% with acquisition enabled, so the recovery comes from verification-gated acquisition rather than from the transferred rules themselves, mirroring the native pattern of Table~\ref{tab:self_evolution}, where frozen $\mathcal{R}_0$ ends at CR 9.63\% against the evolved 5.67\%. Sim-to-real shows slightly smaller residual gaps and lower final call rates than real-to-sim; one plausible factor is its larger source memory (432 against 187 rules before deduplication).

The initial call rate of 86.2\% also shows that most of the source-domain memory is not functionally reusable zero-shot in a new but still driving context. This is consistent with the safety-oriented design: thresholds are calibrated to the source-domain distribution, so under shift the harness re-invokes and re-verifies rather than reusing rules whose calibration may no longer hold. We therefore do not frame these results as evidence of strong zero-shot generalization; the limitation is stated in Section~\ref{sec:limitations}.

\subsection{Latency of the Online-Miss Cycle and Its Scheduling}
\label{app:miss_cycle}

\textbf{What PE counts.} For every method, PE measures the synchronous action-selection computation per step. In the evaluated closed-loop setting, a miss completes proposal, verification, and write-back within the step while simulation time is paused, so the reported PE excludes the acquisition pipeline, namely the LLM call together with the subsequent parsing, calibration, verification, and write-back. Table~\ref{tab:miss_cycle} profiles each stage.

\begin{table}[htbp]
\centering
\fontsize{8.4pt}{10pt}\selectfont
\setlength{\tabcolsep}{6pt}
\begin{tabular}{l|c|c|l}
\toprule
\textbf{Stage} & \textbf{Mean} & \textbf{p95} & \textbf{Deployment scheduling} \\
\midrule
Fallback action for the current step            & 2.6\,ms  & 3.9\,ms  & control path, within budget \\
\midrule
Prompt assembly and state serialization         & 0.9\,ms  & 1.4\,ms  & background \\
LLM rule proposal (GPT-4o API)                  & 1.58\,s  & 2.94\,s  & background \\
Parsing and schema validation                   & 2.1\,ms  & 3.0\,ms  & background \\
Threshold calibration on the sliding window     & 1.3\,ms  & 1.9\,ms  & background \\
Rollout verification at $H{=}5$                 & 1.6\,ms  & 2.2\,ms  & background \\
Neo4j insert and index update                   & 4.7\,ms  & 6.8\,ms  & background \\
\midrule
Full cycle, miss to usable rule                 & 1.59\,s  & 2.96\,s  & background \\
\bottomrule
\end{tabular}
\caption{Latency of the online-miss cycle on Highway. The fallback row sits on the control path and is not part of the full-cycle latency; the acquisition stages below it sum to the full-cycle row.}
\label{tab:miss_cycle}
\end{table}

\textbf{Deployment scheduling.} The cycle takes 1.59\,s mean, dominated by the LLM call; the acquisition stages other than the LLM call total about 11\,ms. In deployment the cycle moves off the critical path by construction, because the planner has a defined behavior without the acquired rule: the best action over the retrieved rules or, with empty retrieval, the safe action that maximizes minimum lookahead TTC (Eq.~\ref{eq:safe_action}) together with a human-takeover alert. The current step executes that fallback while the cycle completes in the background, and the acquired rule serves later encounters. At 5\,Hz the acquisition window spans about eight control cycles, during which the planner behaves as a verified conservative planner. Figure~\ref{fig:schedule} contrasts the evaluated schedule with the deployment schedule.

\begin{figure}[htbp]
\centering
\begin{tikzpicture}[
  font=\scriptsize,
  box/.style={draw, rounded corners=1pt, minimum height=5.5mm, inner sep=2pt, align=center},
  ctrl/.style={box, fill=blue!8},
  acq/.style={box, fill=orange!12},
  arr/.style={-{Latex[length=1.6mm]}, thick},
  lbl/.style={font=\scriptsize\itshape, anchor=east, align=center}
]
% Evaluated schedule
\node[lbl] (l1) at (0,0) {Evaluated\\(sim.\ paused)};
\node[ctrl, right=2mm of l1] (e1) {retrieve\\miss};
\node[acq, right=2mm of e1] (e2) {LLM proposal\\1.58\,s};
\node[acq, right=2mm of e2] (e3) {calibrate +\\verify, 3\,ms};
\node[acq, right=2mm of e3] (e4) {write-back\\5\,ms};
\node[ctrl, right=2mm of e4] (e5) {act $a^\star_t$};
\draw[arr] (e1) -- (e2); \draw[arr] (e2) -- (e3); \draw[arr] (e3) -- (e4); \draw[arr] (e4) -- (e5);
\node[anchor=west, font=\scriptsize] at ([xshift=2mm]e5.east) {PE counts only the control-path boxes};
% Deployment schedule
\node[lbl] (l2) at (0,-1.55) {Deployment\\(control path)};
\node[ctrl, right=2mm of l2] (d1) {retrieve\\miss};
\node[ctrl, right=2mm of d1] (d2) {fallback $a^{\mathrm{safe}}_t$\\2.6\,ms};
\node[ctrl, right=2mm of d2] (d3) {step $t{+}1$\\$\cdots$};
\node[ctrl, right=2mm of d3] (d4) {step $t{+}8$\\memory hit};
\draw[arr] (d1) -- (d2); \draw[arr] (d2) -- (d3); \draw[arr] (d3) -- (d4);
\node[lbl] (l3) at (0,-2.75) {Deployment\\(background)};
\node[acq, right=2mm of l3] (b1) {LLM proposal\\1.58\,s};
\node[acq, right=2mm of b1] (b2) {calibrate +\\verify, 3\,ms};
\node[acq, right=2mm of b2] (b3) {write-back\\5\,ms};
\draw[arr] (b1) -- (b2); \draw[arr] (b2) -- (b3);
\draw[arr, dashed] (d1.south) -- ++(0,-0.45) -| (b1.north);
\draw[arr, dashed] (b3.north) -- ++(0,0.45) -| (d4.south);
\end{tikzpicture}
\caption{Dataflow of the online-miss cycle. Top: in the evaluated closed-loop setting the cycle completes within the step while simulation time is paused, and PE counts only the synchronous control-path computation. Bottom: in deployment the current step executes the verified fallback while the cycle runs in the background (about eight control cycles at 5\,Hz), and the acquired rule serves later encounters.}
\label{fig:schedule}
\end{figure}
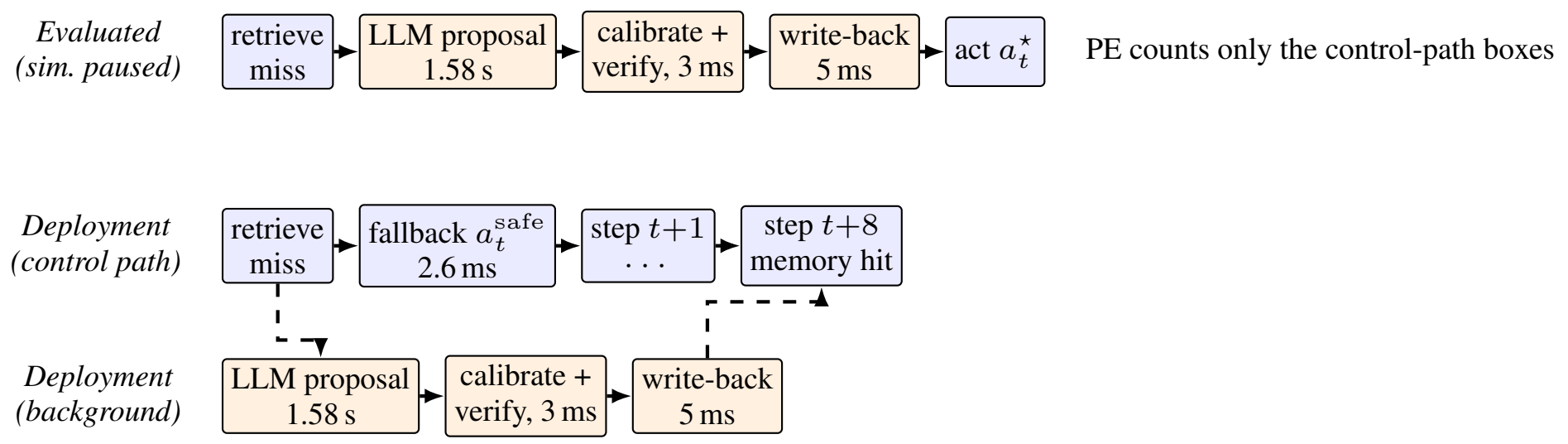

\subsection{Retrieval and Aggregation at Deployment Scale}
\label{app:scalability}

As memory grows, retrieval and multi-rule aggregation must stay within the control window. The stress test runs on Highway and varies only the database size, inflating it with synthetic rules generated as variable and threshold perturbations of compiled rules that preserve the schema and the trigger-degree distribution (Table~\ref{tab:scalability}). Retrieval stays sub-millisecond and grows slowly with memory size, from 0.41\,ms at the paper's 1{,}023 rules to 0.79\,ms at one hundred thousand, consistent with indexed lookup rather than linear scanning. Conflict aggregation does not grow with the database: retrieval returns at most $k{=}10$ rules regardless of memory size, so noisy-OR aggregation (Eq.~\ref{eq:prior}) and the arbitration of Appendix~\ref{app:conflict} cost $O(k)$. On the online planning path, the database-size-dependent cost therefore concentrates in retrieval; index maintenance and write costs sit on the acquisition path, off the control loop.

\begin{table}[htbp]
\centering
\fontsize{8.4pt}{10pt}\selectfont
\setlength{\tabcolsep}{8pt}
\begin{tabular}{l|c|c|c|c}
\toprule
\textbf{Memory size} & \textbf{1,023 (paper)} & \textbf{10,000} & \textbf{50,000} & \textbf{100,000} \\
\midrule
Cypher retrieval (ms) & 0.41 & 0.55 & 0.68 & 0.79 \\
End-to-end PE (ms)    & 2.50 & 2.64 & 2.77 & 2.88 \\
\bottomrule
\end{tabular}
\caption{Retrieval latency and end-to-end PE on Highway as the rule memory is inflated with schema-preserving synthetic rules.}
\label{tab:scalability}
\end{table}
\endgroup

% NeurIPS paper checklist: part of the camera-ready, omitted from the arXiv version.
% \input{checklist.tex}

\end{document}